\documentclass{ecai} 

\usepackage{microtype}
\usepackage{graphicx}
\usepackage{multirow}
\usepackage{makecell}
\usepackage{dsfont}
\usepackage{subcaption}

\usepackage{latexsym}
\usepackage{amssymb}
\usepackage{amsmath}
\usepackage{amsthm}
\usepackage{booktabs}
\usepackage{enumitem}
\usepackage{graphicx}
\usepackage{color}
\usepackage[normalem]{ulem}
\useunder{\uline}{\ul}{}

\usepackage{algorithmic}
\usepackage{algorithm}

\newtheorem{theorem}{Theorem}
\newtheorem{lemma}[theorem]{Lemma}

\usepackage{listings}
\definecolor{codegreen}{rgb}{0,0.6,0}
\definecolor{codegray}{rgb}{0.5,0.5,0.5}
\definecolor{codepurple}{rgb}{0.58,0,0.82}
\definecolor{backcolour}{rgb}{0.95,0.95,0.92}

\lstdefinestyle{mystyle}{
    backgroundcolor=\color{backcolour},   
    commentstyle=\color{codegreen},
    keywordstyle=\color{magenta},
    numberstyle=\tiny\color{codegray},
    stringstyle=\color{codepurple},
    basicstyle=\ttfamily\footnotesize,
    breakatwhitespace=false,         
    breaklines=true,                 
    captionpos=b,                    
    keepspaces=false,                 
    numbers=left,                    
    numbersep=2pt,                  
    showspaces=false,                
    showstringspaces=false,
    showtabs=false,                  
    tabsize=2
}
\newcommand{\BibTeX}{B\kern-.05em{\sc i\kern-.025em b}\kern-.08em\TeX}

\begin{document}


\begin{frontmatter}


\paperid{3548} 


\title{DmC: Nearest Neighbor Guidance Diffusion Model for Offline Cross-domain Reinforcement Learning}


\author[A]{\fnms{Linh}~\snm{Le Pham Van}\thanks{Corresponding Author. Email: l.le@deakin.edu.au}}
\author[A]{\fnms{Minh Hoang}~\snm{Nguyen}}
\author[A]{\fnms{Duc}~\snm{Kieu}}
\author[A]{\fnms{Hung}~\snm{Le}}
\author[B]{\fnms{Hung}~\snm{The Tran}}
\author[A]{\fnms{Sunil}~\snm{Gupta}}

\address[A]{Applied Artificial Intelligence Initiative, Deakin University}
\address[B]{AI Center, VNPT Media, Vietnam}


\begin{abstract}
Cross-domain offline reinforcement learning (RL) seeks to enhance sample efficiency in offline RL by utilizing additional offline source datasets. A key challenge is to identify and utilize source samples that are most relevant to the target domain. Existing approaches address this challenge by measuring domain gaps through domain classifiers, target transition dynamics modeling, or mutual information estimation using contrastive loss. However, these methods often require large target datasets, which is impractical in many real-world scenarios. In this work, we address cross-domain offline RL under a limited target data setting, identifying two primary challenges: (1) Dataset imbalance, which is caused by large source and small target datasets and leads to overfitting in neural network-based domain gap estimators, resulting in uninformative measurements; and (2) Partial domain overlap, where only a subset of the source data is closely aligned with the target domain. To overcome these issues, we propose DmC, a novel framework for cross-domain offline RL with limited target samples. Specifically, DmC utilizes $k$-nearest neighbor ($k$-NN) based estimation to measure domain proximity without neural network training, effectively mitigating overfitting. Then, by utilizing this domain proximity, we introduce a nearest-neighbor-guided diffusion model to generate additional source samples that are better aligned with the target domain, thus enhancing policy learning with more effective source samples. Through theoretical analysis and extensive experiments in diverse MuJoCo environments, we demonstrate that DmC significantly outperforms state-of-the-art cross-domain offline RL methods, achieving substantial performance gains.
\end{abstract}

\end{frontmatter}


\section{Introduction}
Reinforcement Learning (RL) has demonstrated its ability to solve complex real-world problems \cite{mnih2015human, schrittwieser2020mastering}. However, RL typically requires extensive trial-and-error interactions with the environment, which can be infeasible in scenarios where data collection is costly or safety is a concern, such as autonomous driving or healthcare. A common solution is to train policies in a safer, faster source environment (e.g., a simulator) while leveraging a limited amount of real-world target data. This paradigm is known as cross-domain RL \cite{eysenbachoff, xu2023cross, lyu2024odrlabenchmark}.

Previous research has tackled cross-domain RL in various settings, including online, where both domains are online \cite{eysenbachoff, lyucross}, or offline target with online source (hybrid) \cite{niu2022trust, niu2023h2o,van2025hybrid}. In this work, we focus on the cross-domain offline setting \cite{liu2022dara, liu2024beyond, xue2024state}, where both source and target domains are offline. This setting is crucial for enhancing the sample efficiency of offline RL methods \cite{kumar2020conservative, kostrikovoffline,nguyen2025beyond}.
A key idea is to leverage source samples close to the target domain to enhance policy learning. Thus, a core challenge lies in accurately quantifying the domain (dynamics) gap between two domains.
Existing works on cross-domain offline reinforcement learning (RL) have introduced various methods to measure the dynamics gap between domains, such as training domain classifiers \cite{liu2022dara}, using conditional variational autoencoder (CVAE) models to approximate target dynamics \cite{liu2024beyond}, and estimating mutual information \cite{wencontrastive}. However, these approaches typically rely on large target datasets, which are impractical in many real-world applications, such as healthcare. To address this limitation, we focus on cross-domain offline RL in the limited target data setting, which brings several challenges that require novel solutions.

We begin by carefully analyzing the challenges in the cross-domain offline RL with \emph{limited} target data setting. 
Specifically, we identify two major challenges (as illustrated in Figure \ref{fig:cls_error}, Figure \ref{fig:log_distance}): (1) Source-target dataset imbalance, which can cause neural network-based domain gap estimators to overfit or be biased toward a large amount of source samples, leading to uninformative dynamics gap measurements; and (2) Partial domain overlap, where only a partial subset of the source data is closely aligned with the target domain.
To address the first challenge, we propose a novel estimation to quantify the proximity of source samples to the target domain via $k$-nearest neighbor ($k$-NN) based estimation. This method avoids training neural networks, thereby reducing the risk of overfitting.
To tackle the second challenge, we use the $k$-NN proximity scores to guide a diffusion model \cite{janner2022planning, lu2023synthetic} that generates additional data samples closer to the target domain, enhancing the quality of policy learning. The core idea of our approach is to leverage the $k$-NN estimator both as a measure of cross-domain dynamics gap and as a guiding signal for data generation. By combining accurate estimation with targeted sample generation, our method offers a complete solution to these challenges.
We name our approach Nearest Neighbor Guidance \textbf{D}iffusion \textbf{m}odel for Offline \textbf{C}ross-Domain Offline RL (DmC). In addition to our methodological contributions, we provide a theoretical analysis of DmC and empirically validate its effectiveness across various Gym-MuJoCo environments \cite{todorov2012mujoco, brockman2016openai}. Our results demonstrate that DmC significantly outperforms state-of-the-art cross-domain offline RL methods, achieving substantial performance improvements.
\section{Related Works}
\subsection{Cross-Domain Offline RL}
Cross-domain RL \cite{niu2024comprehensive,niu2024xted} aims to improve sample efficiency in the target domain by leveraging data from additional source environments. Several approaches have been proposed to address this challenge, including system identification \cite{werbos1989neural, zhu2018fast, chebotar2019closing}, domain randomization \cite{sadeghi2017cad2rl, tobin2017domain, peng2018sim}, and meta-RL \cite{finn2017model, nagabandi2018learning, wu2023zero}. However, these methods often require environment models or domain knowledge to carefully select randomized parameters. Recently, several methods have attempted to measure dynamics discrepancy for various settings, including purely online \cite{eysenbachoff,le2024policy,xu2023cross, guo2024off}, purely offline \cite{liu2024beyond, wencontrastive, xue2024state, chen2024domain,wang2024return, nishimori2024offline}, or a hybrid setting \cite{niu2022trust}. In this work, we focus on the cross-domain \emph{offline} setting,  where both the source and target domains are offline. Previous approaches tackle this problem through reward modification \cite{liu2022dara}, support constraints \cite{liu2024beyond}, or data filtering via mutual information \cite{wencontrastive}.
However, these methods require training neural networks (NNs) to estimate the dynamics gap between two domains, which can be challenging when only limited target data is available. 
To overcome this limitation, we propose using $k-$NN based estimation to measure the divergence between source and target domains, avoiding the need for NNs training. The concurrent work from \citet{lyu2025crossdomain} introduced OTDF, which leverages optimal transport (OT) to address neural network training challenges. However, OTDF involves solving a computationally expensive OT problem between source and target datasets. In contrast, $k$-NN estimation offers a more efficient and scalable solution. Moreover, unlike prior works, including OTDF, our approach generates more samples close to the target domain, thus bringing more sample efficiency.
We employ a novel guided-diffusion model to augment the source dataset with samples close to the target domain. These strategies enable our method to perform effectively in limited target data settings.
\subsection{Diffusion Model in RL}
Diffusion models \cite{ho2020denoising, songscore} have been successfully applied as policy models \cite{kang2023efficient, wangdiffusion}, and planners \cite{janner2022planning, liang2023adaptdiffuser, liefficient}, showcasing their effectiveness across various RL tasks, including offline RL \cite{lu2023synthetic, wang2009divergence}, multi-task learning \cite{he2023diffusion}, and meta RL \cite{ni2023metadiffuser}.
Recently, diffusion models have been explored for data augmentation in offline reinforcement learning. Specifically, GTA \cite{leegta} generates high-reward samples, while DiffStitch \cite{li2024diffstitch} constructs stitching trajectories that connect low-reward and high-reward segments. However, in cross-domain settings, high-reward trajectories from the source domain do not necessarily yield high rewards in the target domain due to mismatched dynamics. In contrast to these approaches, our method focuses on generating transitions that are well-aligned with the target domain. To achieve this, we guide a diffusion model using a novel proximity score derived from a non-parametric $k$-NN estimator, which effectively quantifies how similar each source sample is to the target domain, making it different from prior approaches. 

\subsection{Nearest Neighbor in RL}
The nearest neighbor approach has been applied and tailored to various RL problems, such as enforcing policy constraints in offline settings \cite{ran2023policy}, quantifying uncertainty \cite{qiao2024sumo}, performing MixUp-based data augmentation \cite{sander2022neighborhood}, and imitation learning \cite{lyuseabo}.  In contrast, our work studies the cross-domain RL with limited target samples. We propose using a $k$-NN estimator to measure the dynamics gap between the source and target domains. This estimation is used to downweight source samples that are dissimilar to the target domain during policy training. Furthermore, we leverage $k$-NN estimation to guide sampling from a diffusion model, enabling the generation of additional source data closely aligned with the target domain.

\section{Preliminaries}


\subsection{Reinforcement Learning}
We first introduce Markov Decision Processes (MDP) which is denoted as $\mathcal{M} = (\mathcal{S}, \mathcal{A}, \gamma, r, d_0, P)$, where $\mathcal{S}, \mathcal{A}$ are the state and action spaces. The parameter $\gamma \in (0, 1)$ is the discounted factor, $r: \mathcal{S} \times \mathcal{A} \rightarrow \mathbb{R}$ is the reward function, $d_0$ is the initial state distribution and $P$ is the transition dynamics. We assume the rewards are bounded, i.e. $|r(s, a)| \leq r_{\text{max}}, \forall s, a$. We denote a policy $\pi : \mathcal{S} \rightarrow \Delta(\mathcal{A})$ as a map from state space $\mathcal{S}$ to a probability distribution over action space $\mathcal{A}$. Given a policy $\pi$ and a transition dynamics (model) $P$, we denote discounted state-action occupancy as $d^\pi_P(s, a) = (1 - \gamma)\mathbb{E}_{\pi, P}\left[\sum^\infty_{t=0}\gamma^t\mathds{1}(s_t=s, a_t=a)\right]$. We define state-action value function as $Q^{\pi}_{P}(s, a) = \mathbb{E}_{\pi, P}\left[\sum^\infty_{t=0}\gamma^t r(s_t, a_t)| s_0=s, a_0=a\right]$, and the value function as $V^{\pi}_{P}(s) = \mathbb{E}_{a\sim \pi}\left[Q^{\pi}_{P}(s, a)\right]$. The objective of RL is to find a policy that maximizes the expected return $J^P(\pi) = \mathbb{E}_{s\sim d_0}\left[V^{\pi}_{P}(s)\right]$.


We address the cross-domain offline setting, where the source domain is defined as $\mathcal{M}_{src} = \left(\mathcal{S}, \mathcal{A}, \gamma, r, d_0, P_{src}\right)$ and the target domain as $\mathcal{M}_{tar} = \left(\mathcal{S}, \mathcal{A}, \gamma, r, d_0, P_{tar}\right)$. We assume that the two domains share the same state space $\mathcal{S}$, action space $\mathcal{A}$, discount factor $\gamma$, reward function $r$, and initial state distribution $d_0$, differing only in their dynamics models.
In this setting, we have access to offline datasets collected from both domains: $D_{src}$ from the source domain and $D_{tar}$ from the target domain. The objective is to utilize the additional source dataset $D_{src}$ to improve policy learning for the target domain, using the target dataset $D_{tar}$.
We further denote the empirical MDP induced by a dataset $D$ as $\widehat{\mathcal{M}} = \left(\mathcal{S}, \mathcal{A}, r, d_0, \gamma, \widehat{P}\right)$, where $\widehat{P}$ represents the empirical transition dynamics derived from $D$. In the cross-domain offline setting, this results in two empirical MDPs: $\widehat{\mathcal{M}}_{src}$ with transitions $\widehat{P}_{src}$ and $\widehat{\mathcal{M}}_{tar}$ with transitions $\widehat{P}_{tar}$.


\subsection{Diffusion Model}
Diffusion models \cite{ho2020denoising} are a class of generative models that learn to denoise and generate data from noise iteratively. 
Specifically, diffusion models corrupt clean data samples $x^{0}\sim q\left(x^{0}\right)$ by progressively adding Gaussian noise, following a Markov process \footnote[1]{As both the diffusion model and RL involve time steps, we use \emph{superscripts} to denote diffusion steps and \emph{subscripts} for RL time steps.}:
\begin{equation}
\resizebox{0.85\columnwidth}{!}{
    $
\begin{aligned}
    q\left(x^{1:T}\mid x^{0}\right)	&=\prod_{t=1}^{T}q\left(x^{t}\mid x^{t-1}\right)
    =\prod_{t=1}^{T}\mathcal{N}\left(\sqrt{\alpha^{t}}x^{t-1},\beta^{t}\mathrm{I}\right)
    \\
    q\left(x^{t}\mid x^{0}\right)&=\mathcal{N}\left(\sqrt{\overline{\alpha}^{t}}x^{0},\left(1-\overline{\alpha}^{t}\right)\mathrm{I}\right),\; \text{with: }\overline{\alpha}^{t}=\prod_{i=1}^{t}\alpha^{i}
\end{aligned}
$}
\end{equation}
where $\alpha^t,\beta^{t}$ represents the noise schedule, chosen such that $q\left(x^{T}\mid x^{0}\right)\approx\mathcal{N}(0,\mathrm{I})$. 
Diffusion models then learn to reverse the corruption process for generating data, where the reverse transition $q\left(x^{t-1}\mid x^{t}\right)$ is approximated by a parameterized model $p_{\theta}\left(x^{t-1}\mid x^{t}\right)=\mathcal{N}\left(\mu_{\theta}\left(x^{t}\right),\sigma_{t}\mathrm{I}\right)$. 
The parameterized model is trained by minimizing KL with $q(x^{t-1}|x^t,x^0)$, which is:
\begin{equation}
\resizebox{0.8\columnwidth}{!}{
    $
\begin{aligned}
    q\left(x^{t-1}\mid x^{t},x^{0}\right)=\mathcal{N}\left(\frac{1}{\sqrt{\alpha^{t}}}\left(x^{t}-\frac{1-\alpha^{t}}{\sqrt{1-\overline{\alpha}^{t}}}\epsilon\right),\beta^{t}\mathrm{I}\right)
\end{aligned}
$}
\end{equation}
Here, $\epsilon$ represents the Gaussian noise added to $x^0$ to form $x^t$. Instead of directly matching the mean of $q\left(x^{t-1}\mid x^{t},x^{0}\right)$ with the mean of $p_{\theta}\left(x^{t-1}\mid x^{t}\right)$, i.e., $\mu_{\theta}\left(x^{t}\right)$, a noise network $\epsilon_\theta$ is adopted to estimate $\epsilon$ from $x^t$. This network is trained by minimizing the regression loss:
\begin{equation}
\resizebox{0.8\columnwidth}{!}{
    $
   \min_\theta \mathbb{E}_{x^{0},\epsilon,t}\left[\left\Vert \epsilon-\epsilon_{\theta}\left(\sqrt{\overline{\alpha}^{t}}x^{0}+\sqrt{1-\overline{\alpha}^{t}}\epsilon, t\right)\right\Vert ^{2}\right]
   \label{eq:noise_matching}
   $}
\end{equation}
where $x^0\sim q(x^0)$, $\epsilon\sim\mathcal{N}(0,\mathrm{I})$, and $t\sim\mathcal{U}[0,T]$.

Guided sampling from diffusion models enables generating data conditioned on attributes $y$, effectively modeling $p(x^0|y)$. In our setting, $y$ represents a score function quantifying how closely generated samples align with the target domain. 
Rather than explicitly training a separate score estimator $p(y|x^t)$, which is challenging due to the inherent noise present in $x^t$, we adopt the classifier-free guidance (CFG)\cite{ho2022classifier}. CFG trains a conditional noise model $\epsilon_{\theta'}(x^t, y, t)$ and combine with unconditional counterpart $\epsilon_\theta(x^t, t)$ during sampling:
\begin{equation}
    \hat{\epsilon}_w(x^t, y, t) = w \epsilon_{\theta'}(x^t, y, t) + (1 - w)\epsilon_{\theta}(x^t, t),
    \label{eq:guided_sampling}
\end{equation}
where $w$ is the guidance coefficient that controls the strength of the conditioning. In practice, both conditional and unconditional noise estimators share identical parameters ($\theta' = \theta$), and the unconditional model is implemented by setting $y$ to an empty value, i.e. $y = \emptyset$.
\begin{figure}[!htbp]
    \centering
    \includegraphics[width=0.55\linewidth]{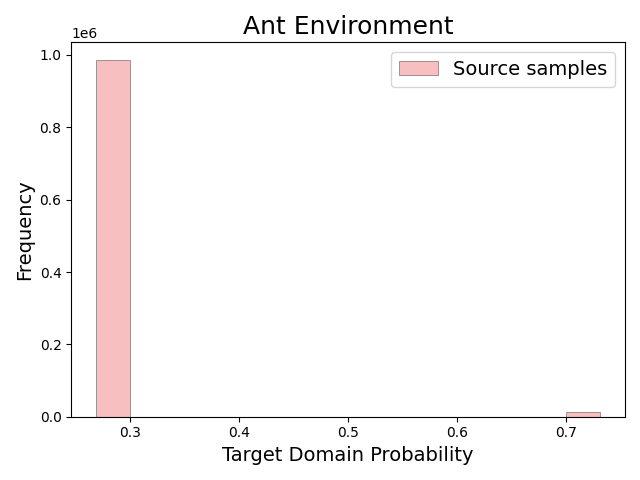}
    \caption{Histogram of target-domain probabilities predicted for source samples in the Ant environment using the pretrained domain classifier (DARA). The classifier concentrated around limited probabilities, offering little information about the domain gaps between the two domains.}
    \label{fig:cls_error}
\end{figure}
\section{Cross-Domain Offline RL with Limited Target Samples}\label{sec:cross_domain_discussion}
In this section, we present the major challenges in cross-domain offline RL with limited target samples. Due to space limits, we defer more detailed discussions in Appendix 2.
\subsection{Datasets Imbalanced Problem}\label{sec:imbalance}
In cross-domain offline RL, the key idea is to leverage source samples that are close to the target domain to improve policy learning and enhance sample efficiency. A central challenge in this setting is to accurately quantify the domain gap between the source and target domains.
Accurate measurement of this gap enables subsequent strategies to leverage source samples to enhance policy learning for the target domain.
Prior works have proposed various approaches for this purpose, including training domain classifiers \cite{liu2022dara}, or estimating target models \cite{liu2024beyond}. These gap estimates are then used in different ways, for instance, as reward penalties that constrain the policy to remain in shared regions across domains \cite{liu2022dara}, or to downweight the value function on source samples that diverge from the target domain \cite{liu2024beyond}.
However, these methods often rely on training parametric models such as neural networks, which are prone to overfitting in scenarios with limited target data and significant dataset imbalances. 
To demonstrate potential overfitting issues, we train the domain classifiers from DARA \cite{liu2022dara} using offline datasets from the Ant environment in D4RL \cite{fu2020d4rl}. As shown in Figure~\ref{fig:cls_error}, the histogram of predicted target-domain probabilities for source samples reveals severe overfitting. The domain classifier produces nearly similar predictions across all source samples, failing to provide meaningful discrimination or insight into domain gaps (as shown in Figure 1 Appendix 2, the reward penalties for source samples are near zero).
In addition to the overfitting issues, the imbalance between source and target data can bias the learned policy, potentially causing it to overfit the source data during training.
\begin{figure}[!htbp]
    \centering
    \includegraphics[width=0.50\linewidth]{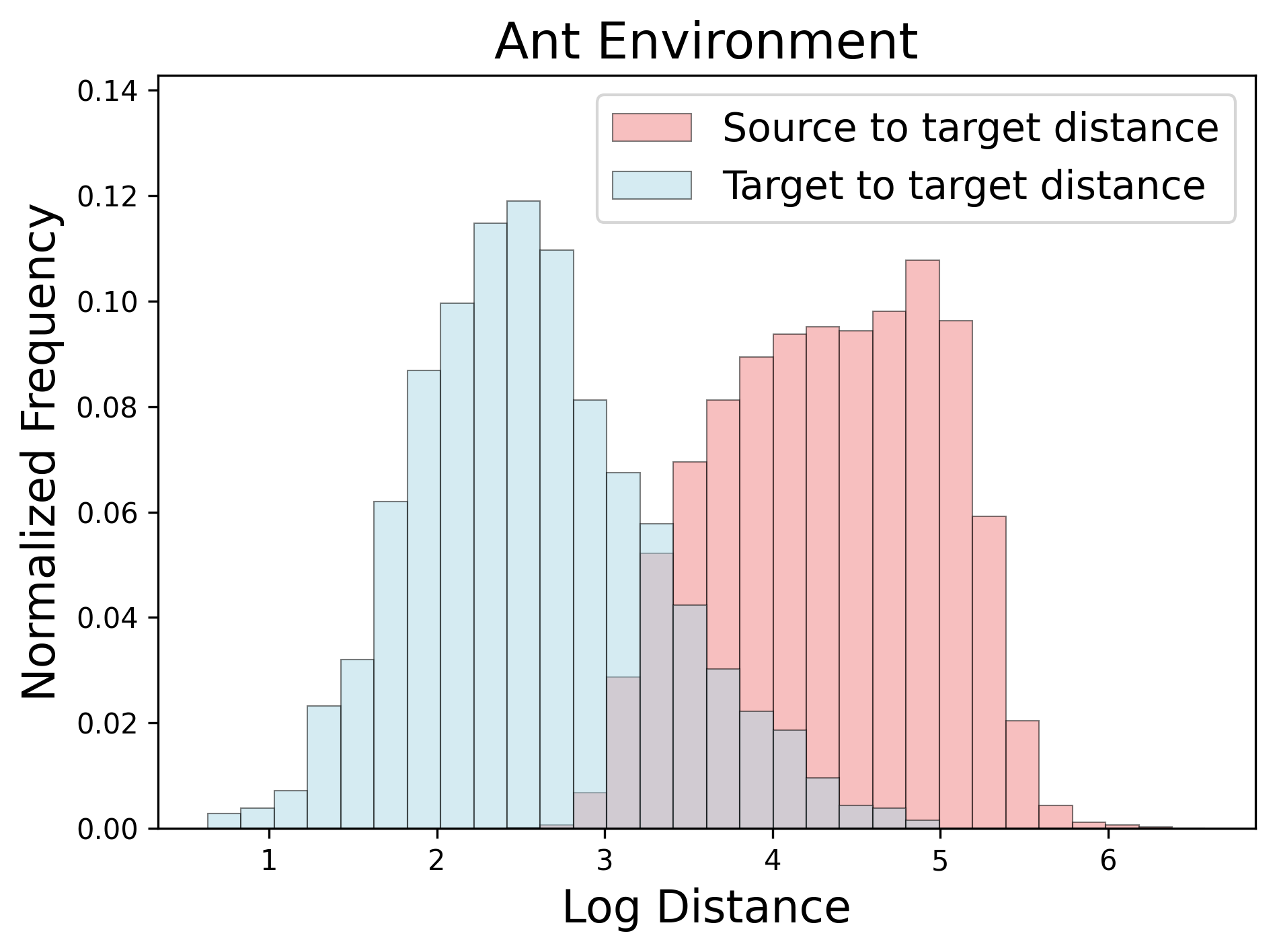}
    \caption{Nearest Neighbor Distance histograms. Blue shows the distance of source samples to their nearest target samples. Pink shows the distance of target samples to their nearest target samples.}
    \label{fig:log_distance}
\end{figure}
\subsection{Partial Overlapping between Domains}
Leveraging source samples is critical for enhancing policy learning and improving sample efficiency in cross-domain offline RL. However, discrepancies in dynamics between the source and target domains, along with potential policy shifts during data collection processes, not all source data is close to the target data. To investigate how the source data is close to the target data, we compute the distances from each source transition $(s, a, s')_{src}$ to its nearest transition in the target dataset $D_{tar}$. Further, to provide a reference, we also compute the distances from each target transition $(s, a, s')_{tar}$ to its nearest neighbor within $D_{tar}$. The histograms in Figure \ref{fig:log_distance} reveal only partial overlap between two histograms, highlighting that only a subset of the source dataset is beneficial for target policy learning. Previous methods \cite{liu2024beyond, wencontrastive} partially addressed this by filtering out source samples far from the target domain.
This raises an important question: ``Can we improve sample efficiency in cross-domain offline RL by generating additional source data that closely aligns with the target domain?"

\section{Diffusion-Guided Cross-Domain Offline RL}
\begin{figure*}[t!]
    \centering
    \includegraphics[width=0.5\linewidth]{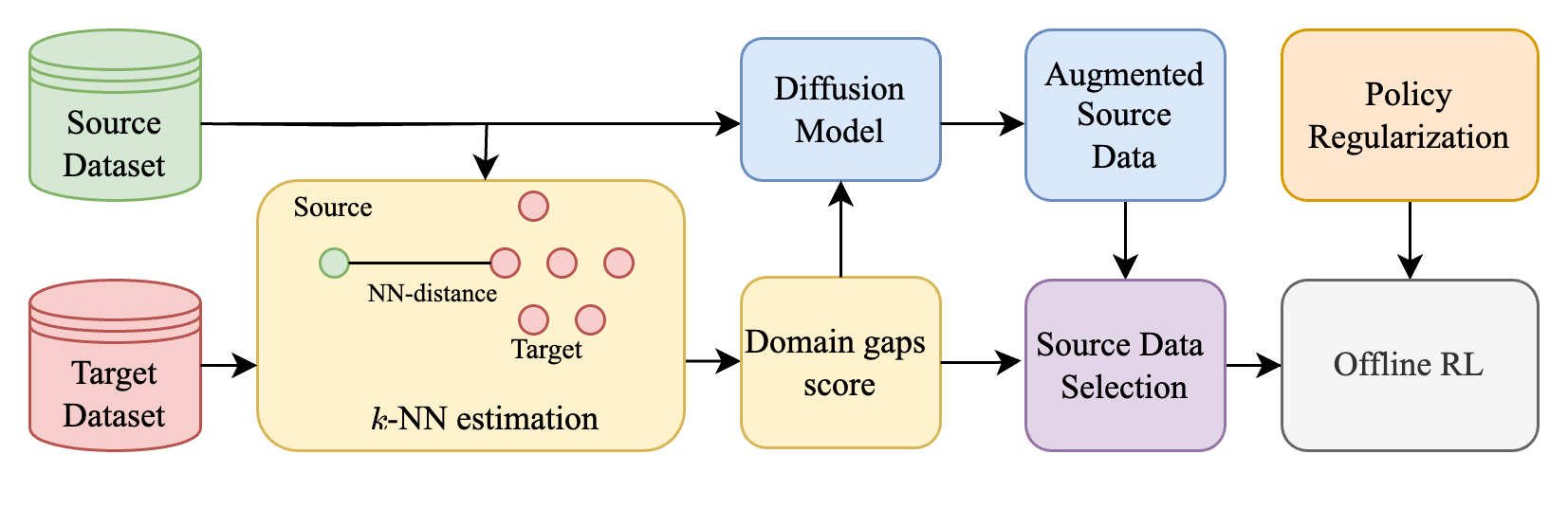}
        \caption{Illustration of our method. First, we use $k$-NN estimation to quantify the domain gap score. Next, we leverage a diffusion model to upsample the source data, generating samples close to the target domain. The datasets are then utilized in an offline RL framework, incorporating a regularization term to ensure the learned policy remains within the support region of the target dataset.}
    \label{fig:main_method}
\end{figure*}
In this section, we introduce our proposed method, illustrated in Figure \ref{fig:main_method}. We first present the $k-$NN estimation for measuring domain gaps between the source and target. We then leverage a guided diffusion model, conditioned on $k-$NN scores, to augment the dataset with source samples that closely resemble the target domain. Next, we formulate a practical algorithm that integrates $k-$NN estimation and guided diffusion for policy learning. Finally, we summarize our approach, DmC, and provide its theoretical analysis.

\subsection{$k$-NN Based Domain Gap Estimation}
Training neural networks often encounters challenges caused by the limited and imbalanced datasets, as presented in Section \ref{sec:imbalance}.
To address this, we propose using $k$-NN estimation as an alternative approach to effectively quantify domain gaps, bypassing the limitations of prior methods.

In the following, we denote $\oplus$ as the concatenation operator, and $x_{tar} = s_{tar} \oplus a_{tar} \oplus s'_{tar}$ and $x_{src} = s_{src} \oplus a_{src} \oplus s'_{src}$, where $s_{tar}, a_{tar}, s'_{tar} \sim D_{tar}$ and $s_{src}, a_{src}, s'_{src} \sim D_{src}$. Thus we have $D_{src} = \{x_{src, i}\}^{i=N}_{i=1}$ and $D_{tar} = \{x_{tar, i}\}^{i = M}_{i = 1}$, where $N, M$ are the sizes of the source dataset and target dataset respectively. We denote $\nu_{k, \text{src}}(i)$ is the Euclidean distance of the $k$th nearest neighbor of $x_{src, i}$ in the source dataset $D_{src}$, and $\nu_{k, \text{tar}}(i)$ is the Euclidean distance of the $k$th nearest neighbor of $x_{src, i}$ in the target dataset $D_{tar}$. Let $\mathcal{B}(x,  R)$ denote a closed ball around $x \in \mathbb{R}^d$ with the radius $R$, and $\mathcal{V}(\mathcal{B}(x, R)) = cR^d$ is its volume, where $c = \frac{\pi^{d/2}}{\Gamma(d/2 + 1)}$ is the volume of a $d$-dimensional unit ball, and $\Gamma$ is the Gamma function. 
Following the $k$-NN based density estimators \cite{wang2009divergence,poczos2011estimation}, given a dataset $D$, we have the following estimator:
\begin{equation}
\resizebox{0.8\columnwidth}{!}{
    $
\begin{aligned}
    P_{D, k} (x^i) = \frac{k/|D|}{\mathcal{V}(\mathcal{B}(x, \nu_{k, D}(x)))} = \frac{k}{|D| c \nu_{k, D}(x)},
\end{aligned}
$}
\end{equation}
where $|D|$ is the size of the dataset $D$ and $\nu_{k, D}(x)$ is the distance between $x$ and its $k$th nearest neighbor in $D$.

We quantify the \emph{domain gaps} between the source and target domains as the $k$-NN estimator for the Kullback-Leibler divergence $\mathcal{D}_{KL}$. Specifically, we measure the estimation of $\mathcal{D}_{KL}$ between source and target domains as follows:
\begin{equation}
\resizebox{0.86\columnwidth}{!}{
    $
    \begin{aligned}
        \widehat{\mathcal{D}}_{KL}(P_{src}||P_{tar}) &= \frac{1}{N}\sum^{N}_{i=1}\left(\log\frac{M\nu^d_{k, \text{tar}}(i)}{(N-1)\nu^d_{k, \text{src}}(i)}\right) \\
        &\propto 1/N \sum^N_{i=1}\left(\log(\nu^d_{k, \text{tar}}(i)) - \log(\nu^d_{k, \text{src}}(i))\right)
    \end{aligned}
    \label{eq:kl_knn}
    $}
\end{equation}
Thus, given a particular source data $x_{src, i} \in D_{src}$, we consider the discrepancy of it to the target dataset as follows:
\begin{equation}
\label{eq:knn_score}
\resizebox{0.9\columnwidth}{!}{
    $
    \begin{aligned}
        &\rho_{k}(x_{src, i})  &= \log (\|x_{src, i} - x_{tar,i}^{k}\|_2) -  \log \left(\left\|x_{src, i} - x_{src,i}^k\right\|_2\right),
    \end{aligned}
    $}
\end{equation}
where $x_{src,i}^k$ and $x_{tar,i}^k$ are the $k$ nearest neighbors of $x_{src, i}$ in the source dataset $D_{src}$ and the target dataset $D_{tar}$ respectively. 
Intuitively, $\rho_{k}(x_{src, i})$ quantifies the discrepancy of the source sample $x_{src, i}$ to the target dataset, and is small if the distance from $x_{src, i}$ to its $k$-nearest neighbors in the target dataset is smaller than the distance to its $k$-nearest neighbors in the source dataset. The critical advantage of $k$-NN estimation lies in its consistency and simplicity: it is stable and avoids the need for neural network training, thus mitigating the risk of overfitting when estimating the domain gap. Furthermore, it is computationally efficient and easy to implement, leveraging the KD-tree data structure. In practice, we use FAISS library \cite{douze2024faiss} in our implementation and compute all scores for 1 million source samples with 5000 target samples within 1 minute.

\subsection{Nearest-Neighbor Guided Diffusion Model}
To overcome the partial overlapping challenge, and bring more sample efficiency for cross-domain offline RL, we propose to upsample the source dataset with generated samples close to the target domain using the diffusion model. To ensure the generated source data is close to the target domain, we opt to use the classifier-free guidance diffusion model with scores computed leveraging the $k$-NN estimation. 

Given the source dataset $D_{src}$, for each source transition $x_{src, i} = (s_{src}, a_{src}, s'_{src})_i$ in $D_{src}$, we leverage the $k$-NN estimation and compute its $\rho_k(x_{src, i})$. The $k$-NN estimation provides a reliable measurement of how close a source sample is to the target domain. Notably, we train a diffusion model on the source dataset $D_{src}$ and the corresponding $k$-NN estimation scores as the conditional context $y$. We adopt the design of EDM \cite{karras2022elucidating} for our diffusion model, as it has demonstrated superior empirical performance in data modeling compared to earlier designs~\cite{ho2020denoising,songscore}. Specifically, we employ a fixed noise schedule $\sigma_{\text{max}}=\sigma^{T}>\dots>\sigma^{1}>\sigma^{0}=0$ and diffuse clean data samples using the transition $q(x^t|x_{src, i})=\mathcal{N}\left(x_{src, i}, (\sigma^t)^2\mathrm{I}\right)$. Then, we train a denoiser $G_\theta(\cdot)$ to directly predict the clean samples using classifier-free training \cite{ho2022classifier}:
\begin{equation}
\resizebox{0.91\columnwidth}{!}{
    $
\begin{aligned}
   \min_\theta \mathbb{E}\left[\left\Vert x_{src, i}-G_{\theta}\left(x_{src, i}+\sigma^{t}\epsilon,m\cdot\rho_k(x_{src, i}),\sigma^{t}\right)\right\Vert ^{2}\right].
\end{aligned} $
}
\label{eq:diffusion_ob}
\end{equation}
where $x_{src, i}\sim D_{src}$, $\epsilon\sim\mathcal{N}(0,\mathrm{I})$, and $t\sim\mathcal{U}[0,T]$. The score $\rho_k(x_{src, i})$ is randomly masked during training via the null token $m$ sampled from the Bernoulli distribution with $p\left(m=\emptyset\right)=0.25$. Notably, the trained denoiser's output can be reformulated into the noise model's objective as $\epsilon=\frac{G_{\theta}\left(x_{src, i}+\sigma^{t}\epsilon,\sigma^{t}\right)-x^t}{\sigma^t}$. Therefore, we can use the trained denoiser to perform guided sampling as introduced in Eq. (\ref{eq:guided_sampling}) for the conditional generation.

The training process for the diffusion model is done before the training of the RL policy. Once the diffusion model is trained, we generate source samples that are close to the target domain by setting the value of the conditional context $y$ to be higher than the top $\kappa\%$ of the source sample scores in source dataset $D_{src}$. Specifically, we uniformly sample a value $\chi$ from the range $[\kappa, 100]$ and set $y$ to the $\chi$-th quantile of the source dataset’s score distribution. Finally, generated samples are combined with source dataset for later policy learning.
\subsection{Algorithm}
Based on the $k$-NN estimation as the domain gaps measurement and the guided-diffusion model for upsampling source dataset with generated samples close to the target domain, we obtain a practical policy adaptation algorithm, named DmC. We summarize our proposed algorithm in Algorithm \ref{alg:DmC}. In practice, we normalize the value of $\rho_k$ to $[0, 1]$. Specifically, we first subtract each value with the minimum value of $\rho_k$ in the source dataset $D_{src}$ to adjust the value to the range $[0, +\infty]$, and then scale it as follows: 
\begin{equation}\label{eq:score_diffusion}
    w_{k}(x_{src, i}) = 1/\left(1 + \hat{\rho}_{k}(x_{src, i})\right),
\end{equation}
where the input $\hat{\rho}_{k}(x_{src, i})$ is the adjusted non-negative value of $\rho_{k}(x_{src, i})$. Intuitively, the larger the $w_{k}$ value, the closer is the source sample $x_{src, i}$ to the target domain. We select the source data close to the target domain to reduce the dynamics gap while training. Specifically, we formulate our objective function for training the value function $Q_{\phi}(s, a)$ as follows:
\begin{equation}
\label{eq:Q}
\resizebox{\columnwidth}{!}{
    $
    \begin{aligned}
        \mathcal{L}_{Q} = \mathbb{E}_{D_{tar}}\left[(Q_\phi - \mathcal{T}Q_\phi)^2\right] 
        + \mathbb{E}_{D_{src}}\left[\omega(s, a, s')(Q_\phi - \mathcal{T}Q_\phi)^2\right].
    \end{aligned}
    $}
\end{equation}
where $\mathcal{T}$ is the Bellman operator, $\omega(s, a, s') := w_k \mathds{1}(w_k \geq w_{k,\xi\%})$, $\mathds{1}$ is the indicator function, and $w_{k, \xi\%}$ denotes the top $\xi$-quantile score used for source data selection. Eq (\ref{eq:Q}) ensures the policy adaptively emphasizes source data close to the target domain, improving adaptation performance. As mentioned, the imbalanced dataset could bias the policy to source samples. Thus, we employ a policy regularization to ensure the learned policy is close to the support areas of the target dataset. Similar to \citet{wu2022supported}, we learn a CVAE, denoted as $\hat{\pi}_{tar}^b(a|s)$,  to model the behavior target policy. Thus, we optimize the policy as follows:
\begin{equation}
\label{eq:pi}
    \mathcal{L}_{\pi}^{\text{reg}} = \mathcal{L}_{\pi} 
- \lambda \mathbb{E}_{s \sim D_{src}\bigcup D_{tar}}\left[\log \hat{\pi}_{tar}^b\left(\pi(.|s)|s\right) \right].
\end{equation}
where $\mathcal{L}_\pi$ is the policy loss of the offline RL method, $\lambda$ is the coefficient controlling the strength of the additional policy regularization. We note that DmC can be integrated with any offline RL method. In our implementation, we use IQL \cite{kostrikovoffline} as our backbone of DmC.

\subsection{Theoretical Analysis}\label{sec:theoretical_analysis}

We provide a theoretical guarantee for using the source dataset to improve the performance in the target domain under the cross-domain offline RL setting. Specifically, we have the following performance bound for any policy $\pi$:

\begin{theorem}\label{theorem:performance_bound} Denote $D_{src}$ as the offline source dataset from source domain $\mathcal{M}_{src}$ and $D_{tar}$ as the offline target dataset from target domain $\mathcal{M}_{tar}$. Let the empirical policy in the offline target dataset $D_{tar}$ be $\pi_{D_{tar}} = \frac{\sum_{D_{tar}}\mathds{1}(s, a)}{\sum_{D_{tar}}\mathds{1}(s)}$. Given a policy $\pi$, we have the following:
\begin{equation}\label{eq:performance_bound}
\resizebox{0.8\columnwidth}{!}{
    $
\begin{aligned}
    &J^{P_{tar}}(\pi) - J^{\widehat{P}_{src}}(\pi) \\& \geq - \frac{2r_{\text{max}}\gamma}{(1 - \gamma)^2} \mathbb{E}_{s, a \sim d^{\pi}_{P_{tar}}, s'\sim P_{tar}} \left[ D_{TV}\left(\pi(.|s') || \pi_{D_{tar}}(.|s')\right) \right] \\
    &-\frac{2r_{\text{max}}\gamma}{(1-\gamma)^2}  \mathbb{E}_{s, a\sim d^{\pi_{D_{tar}}}_{P_{tar}}, s'\sim P_{tar}}\left[D_{TV}\left(\pi(.|s') || \pi_{D_{tar}}(.|s')\right)\right] \\
    &- \frac{2r_{\text{max}}\gamma}{(1-\gamma)^2} \mathbb{E}_{s, a\sim d^{\pi_{D_{tar}}}_{P_{tar}}} \left[\sqrt{1/2D_{KL}(P_{tar}(.|s, a) || \widehat{P}_{src}(.|s, a))}\right].
\end{aligned}
$}
\end{equation}
\end{theorem}

We provide detailed proof in Appendix 1. We recall that $J^{P}(\pi)$ is the expected return of $\pi$ in $P$.
The first and the second terms of the divergence in Eq. (\ref{eq:performance_bound}) measure the deviation between the current learned policy and the target behavior policy; the third term measures the dynamics mismatch between the actual target dynamics model and the estimated source dynamics model. Motivated by Theorem \ref{theorem:performance_bound}, our proposed method aims to reduce the third term by \emph{effectively measuring the dynamics gaps} between two domains and \emph{generating more source samples close to the target domain}. Specifically, our $k$-NN estimation $\rho_{k}$ estimates the KL divergence between the source dynamics and the target dynamics as shown in Eq. (\ref{eq:kl_knn}). 
We then train a diffusion model on the source dataset, guided by a correspondence score $w_k$ (Eq. \ref{eq:score_diffusion}). By sampling from this trained guided-diffusion model, we generate source samples \emph{close} to the target domain, thereby directly \emph{reducing the dynamics gap}. We further mitigate this gap during Q-function updates (Eq. \ref{eq:Q}, Figure \ref{fig:main_method}) by selectively utilizing the source samples closest to the target domain. Additionally, our algorithm incorporates a policy regularization to address policy deviation between the learned policy and the target behavior policy. 

\begin{algorithm}[ht]
\caption{Guided-\textbf{D}iffusio\textbf{n} Cross-\textbf{D}omain Offline RL -- DmC}
\label{alg:DmC}
\begin{algorithmic}[1]\label{alg:proposed}
\STATE\textbf{Input:} Offline source $D_{src}$, offline target $D_{tar}$ datasets.  
\STATE \textbf{Initialize:} Value function $Q$, policy $\pi$, diffusion model $G_\theta$, $k$ NN value, coefficient $\lambda$, source score buffer $SB$.

\FOR {each source sample $x_{src, i} \in D_{src}$}
    \STATE Compute its k-NN score $\rho_k(x_{src,i})$ via Eq. (\ref{eq:knn_score}).
    \STATE $SB \leftarrow SB \bigcup \{\rho_k(x_{src,i})\}$. 
\ENDFOR

\STATE Train $G_\theta$ with $D_{src}$ and $SB$ via Eq. (\ref{eq:diffusion_ob}).

\STATE Generate samples from the diffusion model and add to the source dataset $D_{src}$ and source score buffer $SB$.

\FOR {each iteration}
    \STATE Sample a batch $b_{src} = \{s, a, s', r\}$ and the corresponding scores $w_{src}$ from $D_{src}$ and $SB$.
    \STATE Sample a batch $b_{tar} = \{s, a, s', r\}$ from $D_{tar}$.
    \STATE Update $Q$ using $b_{src}$, $w_{src}$ and $b_{tar}$ via Eq. (\ref{eq:Q}).
    \STATE Update $\pi$ using $b_{src}$ and $b_{tar}$ via Eq. (\ref{eq:pi}).
\ENDFOR
\STATE\textbf{return } $Q, \pi$.
\end{algorithmic}
\end{algorithm}
\begin{table*}[t!]
\centering
\caption{Results in gravity shift tasks. We report normalized scores and their standard deviations in the target domain, averaged over five random seeds. The best score is \textbf{bold}. Half=Halfcheetah, Hopp=Hopper, m=medium, me=medium-expert, mr=medium-replay, e=expert.}
\label{tab:main_gravity}
\resizebox{0.65\textwidth}{!}{%
\begin{tabular}{@{}ll|lllllll@{}}
\toprule
\multicolumn{1}{c}{Source} & \multicolumn{1}{c|}{Target} & \multicolumn{1}{c}{IQL}   & \multicolumn{1}{c}{DARA}  & \multicolumn{1}{c}{BOSA}  & \multicolumn{1}{c}{SRPO}  & \multicolumn{1}{c}{IGDF}  & \multicolumn{1}{c}{OTDF}   & \multicolumn{1}{c}{DmC}             \\ \midrule
Ant-m                      & m                           & 10.2±1.8                  & 9.4±0.9                   & 12.4±2.0                  & 11.7±1.0                  & 11.3±1.3                  & 45.1±12.4                  & \textbf{56.9±2.2}                   \\
Ant-m                      & me                          & 9.4±1.2                   & 10.0±0.9                  & 11.6±1.3                  & 10.2±1.2                  & 9.4±1.4                   & 33.9±5.4                   & \textbf{47.5±3.9}                   \\
Ant-m                      & e                           & 10.2±0.3                  & 9.8±0.6                   & 11.8±0.4                  & 9.5±0.6                   & 9.7±1.6                   & 33.2±9.0                   & \textbf{36.1±7.8}                   \\
Ant-me                     & m                           & 9.8±2.4                   & 8.1±1.8                   & 8.1±3.0                   & 8.4±2.1                   & 8.9±1.5                   & 18.6±11.9                  & \textbf{55.7±8.1}                   \\
Ant-me                     & me                          & 9.0±0.8                   & 6.4±1.4                   & 6.2±1.5                   & 6.1±3.5                   & 7.2±2.9                   & 34.0±9.4                   & \textbf{46.3±4.9}                   \\
Ant-me                     & e                           & 9.1±2.6                   & 10.4±2.9                  & 4.2±3.9                   & 8.8±1.0                   & 9.2±1.5                   & 23.2±2.9                   & \textbf{42.7±13.0}                  \\
Ant-mr                     & m                           & 18.9±2.6                  & 21.7±2.1                  & 13.9±1.5                  & 18.7±1.7                  & 19.6±1.0                  & 29.6±10.7                  & \textbf{41.8±4.7}                   \\
Ant-mr                     & me                          & 19.1±3.0                  & 18.3±2.1                  & 15.9±2.7                  & 18.7±1.8                  & 20.3±1.6                  & 25.4±2.1                   & \textbf{27.6±0.1}                   \\
Ant-mr                     & e                           & 18.5±0.9                  & 20.0±1.3                  & 14.5±1.7                  & 19.9±2.1                  & 18.8±2.1                  & 24.5±2.8                   & \textbf{28.0±0.3}                   \\
Half-m                     & m                           & 39.6±3.3                  & 41.2±3.9                  & 38.9±4.0                  & 36.9±4.5                  & 36.6±5.5                  & 40.7±7.7                   & \textbf{48.0±0.6}                   \\
Half-m                     & me                          & 39.6±3.7                  & 40.7±2.8                  & 40.4±3.0                  & 40.7±2.3                  & 38.7±6.2                  & 28.6±3.2                   & \textbf{48.9±0.7}                   \\
Half-m                     & e                           & 42.4±3.8                  & 39.8±4.4                  & 40.5±3.9                  & 39.4±1.6                  & 39.6±4.6                  & 36.1±5.3                   & \textbf{48.8±1.0}                   \\
Half-me                    & m                           & 38.6±6.0                  & 37.8±3.3                  & 41.8±5.1                  & 42.5±2.3                  & 37.7±7.3                  & 39.5±3.5                   & \textbf{49.9±1.4}                   \\
Half-me                    & me                          & 39.6±3.0                  & 39.4±4.4                  & 38.7±3.7                  & 43.3±2.7                  & 40.7±3.2                  & 32.4±5.5                   & \textbf{48.1±1.9}                   \\
Half-me                    & e                           & 43.4±0.9                  & 45.3±1.3                  & 39.9±2.7                  & 43.3±3.0                  & 41.1±4.1                  & 26.5±9.1                   & \textbf{51.0±0.8}                   \\
Half-mr                    & m                           & 20.1±5.0                  & 17.6±6.2                  & 20.0±4.9                  & 17.5±5.2                  & 14.4±2.2                  & 21.5±6.5                   & \textbf{32.2±0.8}                   \\
Half-mr                    & me                          & 17.2±1.6                  & 20.2±5.2                  & 16.7±4.2                  & 16.3±1.7                  & 10.0±2.5                  & 14.7±4.1                   & \textbf{30.3±5.1}                   \\
Half-mr                    & e                           & 20.7±5.5                  & 22.4±1.7                  & 15.4±4.2                  & 23.1±4.0                  & 15.3±3.7                  & 11.4±1.9                   & \textbf{32.7±2.6}                   \\
Hopp-m                     & m                           & 11.2±1.1                  & 17.3±3.8                  & 15.2±3.3                  & 12.4±1.0                  & 15.3±3.5                  & \textbf{32.4±8.0}          & 30.6±5.5                            \\
Hopp-m                     & me                          & 14.7±3.6                  & 15.4±2.5                  & 21.1±9.3                  & 14.2±1.8                  & 15.1±3.6                  & 24.2±3.6                   & \textbf{35.7±1.4}                   \\
Hopp-m                     & e                           & 12.5±1.6                  & 19.3±10.5                 & 12.7±1.7                  & 11.8±0.9                  & 14.8±4.0                  & 33.7±7.8                   & \textbf{51.3±10.6}                  \\
Hopp-me                    & m                           & 19.1±6.6                  & 18.5±12.3                 & 15.9±5.9                  & 19.7±8.5                  & 22.3±5.4                  & 26.4±10.1                  & \textbf{52.2±6.6}                   \\
Hopp-me                    & me                          & 16.8±2.7                  & 16.0±6.1                  & 17.3±2.5                  & 15.8±3.3                  & 16.6±7.7                  & 28.3±6.7                   & \textbf{50.4±11.3}                  \\
Hopp-me                    & e                           & 20.9±4.1                  & 23.9±14.8                 & 23.2±7.9                  & 21.4±1.9                  & 26.0±9.2                  & 44.9±10.6                  & \textbf{52.8±10.3}                  \\
Hopp-mr                    & m                           & 13.9±2.9                  & 10.7±4.3                  & 3.3±1.9                   & 14.0±2.6                  & 15.3±4.4                  & 31.1±13.4                  & \textbf{35.4±0.7}                   \\
Hopp-mr                    & me                          & 13.3±6.3                  & 12.5±5.6                  & 4.6±1.7                   & 14.4±4.2                  & 15.4±5.5                  & 24.2±6.1                   & \textbf{41.0±2.0}                   \\
Hopp-mr                    & e                           & 11.0±2.6                  & 14.3±6.0                  & 3.2±0.8                   & 16.4±5.0                  & 16.1±4.0                  & \textbf{31.0±9.8}          & 29.9±7.8                            \\
Walker-m                   & m                           & 28.1±12.9                 & 28.4±13.7                 & 38.0±11.2                 & 21.4±7.0                  & 22.1±8.4                  & 36.6±2.3                   & \textbf{52.5±2.0}                   \\
Walker-m                   & me                          & 35.7±4.7                  & 30.7±9.7                  & 40.9±7.2                  & 34.0±9.9                  & 35.4±9.1                  & 44.8±7.5                   & \textbf{59.2±2.7}                   \\
Walker-m                   & e                           & 37.3±8.0                  & 36.0±7.0                  & 41.3±8.6                  & 39.5±3.8                  & 36.2±13.6                 & 44.0±4.0                   & \textbf{63.8±2.7}                   \\
Walker-me                  & m                           & 39.9±13.1                 & 41.6±13.0                 & 32.3±7.2                  & 46.4±3.5                  & 33.8±3.1                  & 30.2±9.8                   & \textbf{57.5±3.3}                   \\
Walker-me                  & me                          & 49.1±6.9                  & 45.8±9.4                  & 40.1±4.5                  & 36.4±3.4                  & 44.7±2.9                  & 53.3±7.1                   & \textbf{67.8±4.0}                   \\
Walker-me                  & e                           & 40.4±11.9                 & 56.4±3.5                  & 43.7±4.4                  & 45.8±8.0                  & 45.3±10.4                 & 61.1±3.4                   & \textbf{67.1±4.8}                   \\
Walker-mr                  & m                           & 14.6±2.5                  & 14.1±6.1                  & 7.6±5.8                   & 17.9±3.8                  & 11.6±4.6                  & 32.7±7.0                   & \textbf{42.2±5.9}                   \\
Walker-mr                  & me                          & 15.3±1.9                  & 15.9±5.8                  & 4.8±5.8                   & 15.3±4.5                  & 13.9±6.5                  & \textbf{31.6±6.1}          & 31.5±5.2                            \\
Walker-mr                  & e                           & 15.8±7.2                  & 15.7±4.5                  & 7.1±4.6                   & 13.7±8.1                  & 15.2±5.3                  & 31.3±5.3                   & \textbf{39.3±6.2}                   \\ \midrule
Total Score                &                             & \multicolumn{1}{c}{825.0} & \multicolumn{1}{c}{851.0} & \multicolumn{1}{c}{763.2} & \multicolumn{1}{c}{825.5} & \multicolumn{1}{c}{803.6} & \multicolumn{1}{c}{1160.7} & \multicolumn{1}{c}{\textbf{1632.7}} \\ \bottomrule
\end{tabular}%
}
\end{table*}
\section{Experiments}
In this section, we present empirical evaluations of our method, focusing on the following questions: (1) Does DmC improve sample efficiency for the base method and outperform strong baselines in cross-domain offline RL under limited target data settings? (2) How does $k$-NN estimation impact DmC's performance? 
(3) Does $k$-NN Guidance Diffusion Model help reduce the dynamics gap? (4) Does $k$-NN Guidance Diffusion Model improve the policy's performance?
Furthermore, we conduct parameter studies to provide deeper insights into the behavior and effectiveness of DmC. We defer more experimental results in Appendix 5 due to space limits.
\subsection{Datasets and Baselines}
To evaluate the policy adaptation performance, we conducted experiments on four environments from the Gym-Mujoco framework \cite{brockman2016openai, todorov2012mujoco}(Ant, Halfcheetah, Hopper, Walker) and considered the gravity and kinematic shifts as the dynamics shifts between the source and the target domains. Specifically, we modify gravity strength to induce the gravity shift and restrict the rotation range of certain joints to create the kinematic shift. For offline source datasets, we use D4RL \cite{fu2020d4rl} and consider target datasets of varying quality—\emph{medium}, \emph{medium-expert}, and \emph{expert}—following D4RL standards. Each target dataset contains 5000 transitions to enforce a limited target data setting. Please see Appendix 3 for more details about the environment settings. 

We evaluate our method DmC against the state-of-the-art cross-domain offline RL methods: \textbf{DARA} \cite{liu2022dara}, which trains the domain classifiers to measure domain gaps; \textbf{BOSA} \cite{liu2024beyond}, which uses support-constrained regularization to ensure the value function and policy are optimized using in-support samples; \textbf{IGDF} \cite{wencontrastive}, which quantifies domain gaps using a mutual information score function; \textbf{SRPO} \cite{xue2024state} that does reward modification via the stationary state distribution; and \textbf{OTDF} \cite{lyu2025crossdomain} that leverages optimal transport to estimate the domain gaps. We also compare DmC with the state-of-the-art offline RL algorithm, \textbf{IQL} \cite{kostrikovoffline}, trained on a mixture of the source and target datasets. Details of baselines are provided in Appendix 4.
We evaluate the performance on the target domain in offline settings using the normalized score. All methods are trained for 1 million steps across 5 random seeds. Table \ref{tab:main_gravity} shows the performance of DmC and other baselines under the \emph{gravity shift} setting. Due to space constraints, results for the \emph{kinematic shift} setting are included in Table 5 in Appendix 5.
\subsection{Performance Evaluation}
\textbf{Answering question 1):} Table \ref{tab:main_gravity} and Table 5 in Appendix 5 demonstrate that DmC consistently outperforms IQL, winning in \textbf{36} out of 36 tasks under the gravity shift and \textbf{35} out of 36 tasks under the kinematic shift. Notably, DmC achieves a \textbf{97.9\%} improvement in target normalized scores under gravity shift tasks and a \textbf{59.4\%} improvement under kinematic shift tasks.
Compared to other baselines, DmC achieves a significant performance gain. Under the gravity shift, DmC outperforms in \textbf{33} out of 36 tasks, attaining a total normalized score of \textbf{1632.7}, surpassing the second-best baseline, OTDF, by \textbf{40.7\%}. Under the kinematic shift, DmC excels in \textbf{29} out of 36 tasks, achieving a total normalized score of \textbf{1902.2}, compared to 1547.6 for OTDF. These results validate the effectiveness of our approach. 

We observe that cross-domain methods relying on neural network-based dynamics gap estimation (IGDF, BOSA, DARA) perform similarly to IQL across many tasks, indicating their limitations in effective offline policy adaptation. This is likely due to challenges in training neural networks on imbalanced datasets with limited target data, as discussed in Section \ref{sec:cross_domain_discussion} and Appendix 2. In contrast, DmC employs $k-$NN estimation that does not require training neural networks. While OTDF leverages optimal transport to mitigate issues arising from limited target data and shows improvements over other baselines, it does not augment the dataset. DmC further enhances sample efficiency by leveraging a diffusion model to generate target-aligned data, leading to improved performance in the target domain.
\subsection{Ablation Studies}
\subsubsection{Effect of K-NN estimator}
\begin{table}[h]
\centering
\caption{Performance comparison between DmC-K and cross-domain RL baselines utilizing NN-based domain gap estimation. We report the total score overall target tasks within each category (detailed results are provided in Appendix 5 Table 8). We bold the highest scores and underline the second-best. me=medium-expert, mr=medium-replay.}
\label{tab:knn_only}
\resizebox{\columnwidth}{!}{%
\begin{tabular}{lcccc}
\hline
\textbf{Group} & \textbf{DARA}    & \textbf{IGDF}    & \textbf{OTDF}            & \textbf{DmC-K}            \\ \hline
Ant me-*       & 24.9 $\pm$ 2.03  & 25.3 $\pm$ 1.97  & \textbf{75.8 $\pm$ 8.07} & {\ul 61.0 $\pm$ 2.87}     \\
Ant mr-*       & 60.0 $\pm$ 1.83  & 58.7 $\pm$ 1.57  & {\ul 79.5 $\pm$ 5.20}    & \textbf{84.6 $\pm$ 0.81}  \\
Walker me-*    & 143.8 $\pm$ 8.63 & 123.8 $\pm$ 5.47 & {\ul 144.6 $\pm$ 6.77}   & \textbf{187.0 $\pm$ 2.70} \\
Walker mr-*    & 45.7 $\pm$ 5.47  & 40.7 $\pm$ 5.47  & \textbf{95.6 $\pm$ 6.13} & {\ul 93.3 $\pm$ 4.97}     \\ \hline
\end{tabular}
}%
\end{table}
\noindent\textbf{Answering question 2):} 
To assess the effectiveness of $k-$NN estimation in capturing dynamics gaps, we evaluate DmC-K, a variant of our method that relies solely on the $k-$NN component, against offline cross-domain RL baselines. Table~\ref{tab:knn_only} summarizes overall performance scores on the target tasks (with detailed results in Appendix Table 8). 
As shown, DmC-K consistently outperforms methods that rely on neural networks to estimate domain gaps, achieving performance improvements ranging from 50\% to 144\%. Additionally, despite its simplicity, DmC-K performs on par with, or slightly better than, OTDF. These results highlight the effectiveness of the $k$-NN estimator in accurately capturing the dynamics gap between two domains.

\subsubsection{Effect of Guided-Diffusion Model}
\begin{figure}[t]
    \centering
    \includegraphics[width=0.6\linewidth]{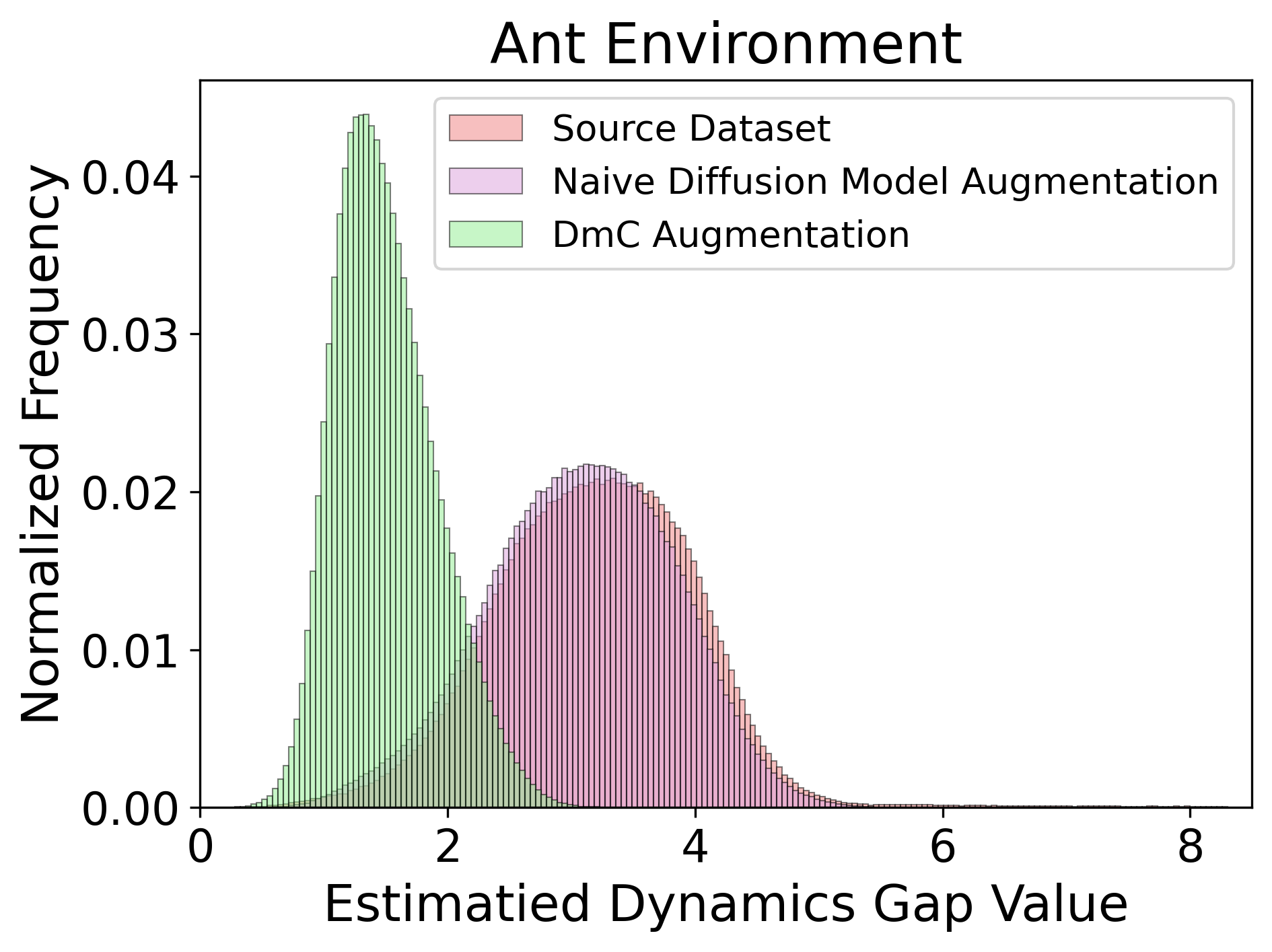}
    \caption{Estimated dynamics gaps between target and different source variants. DmC distribution is concentrated at a lower gap value.}
    \label{fig:kl_hist}
\end{figure}
\noindent\textbf{Answering question 3):} To investigate the effectiveness of the $k$-NN guided diffusion model in reducing the dynamics gap, we estimate the gap for each sample in three datasets: the original source dataset, samples generated by an unconstrained diffusion model, and samples generated by our $k$-NN guided diffusion model. As shown in Figure \ref{fig:kl_hist}, our method significantly reduces the dynamics gap. While the unconstrained diffusion model tends to reproduce the distribution of the source dataset, our $k$-NN guided diffusion model generates samples that are notably closer to the target domain, as evidenced by the lower estimated gaps.

\begin{table}[t]
\centering
\caption{Performance comparison between DmC and its variants on using the diffusion model. We bold the highest scores. m=medium, e=expert, me=medium-expert.}
\label{tab:diffusion_effect}
\resizebox{0.62\columnwidth}{!}{%

\begin{tabular}{@{}llll@{}}
\toprule
Method           & m-m               & m-me              & m-e               \\ \midrule
w/o diffusion    & 23.6±2.9          & 14.4±0.8          & 19.7±1.4          \\
upsample target & 25.4±3.1          & 23.7±3.7          & 24.2±2.7          \\
w/o guidance     & 46.8±11.5         & 31.6±11.5         & 22.4±3.6           \\
w guidance (DmC)       & \textbf{56.9±2.2} & \textbf{47.5±3.9} & \textbf{36.1±7.8} \\ \bottomrule
\end{tabular}
}
\end{table}
\noindent\textbf{Answering question 4):} We compare DmC against three variants: (1) DmC without upsampling via the diffusion model (\emph{w/o diffusion}), (2) DmC where the target dataset is upsampled using a diffusion model (\emph{upsample target}), and (3) DmC with naive diffusion-based generation without guidance (\emph{w/o guidance}). We conduct experiments on Ant \emph{medium} source dataset under gravity shift and present results in Table \ref{tab:diffusion_effect}.
We observe that \emph{upsample target} slightly improves performance over \emph{w/o diffusion}. However, the \emph{w/o guidance} variant, which upsamples the source dataset, achieves better results than \emph{upsample target}. We hypothesize that the limited target data setting hinders the training diffusion model in the target domain, thus affecting adaptation performance. Among all variants, DmC achieves the highest performance, demonstrating the effectiveness of our proposed method of using a guided-diffusion model to upsample the source dataset with generated data close to the target domain.
\subsubsection{DmC with other domain gaps measurement}
\begin{table}[t]
\centering
\caption{Performance comparison between DmC and its variant using classifiers score to measure the domain gaps.}
\label{tab:cls_score}
\resizebox{0.7\columnwidth}{!}{%

\begin{tabular}{@{}llll@{}}
\toprule
Source   & Target & \multicolumn{1}{c}{Classifier score} & \multicolumn{1}{c}{$k$-NN score (DmC)} \\ \midrule
Ant-m    & m      & 31.5±5.3                             & \textbf{56.9±2.2}                      \\
Ant-m    & me     & 15.0±0.8                             & \textbf{47.5±3.9}                      \\
Ant-m    & e      & 19.2±1.9                             & \textbf{36.1±7.8}                      \\
Walker-m & m      & 52.1±3.5                             & \textbf{52.5±2.0}                      \\
Walker-m & me     & \textbf{59.2±2.7}                    & \textbf{52.9±2.7}                      \\
Walker-m & e      & 61.9±2.6                             & \textbf{63.8±2.7}                      \\ \bottomrule
\end{tabular}
}
\end{table}
To further assess the importance of $k-$NN estimation in DmC, we replace the $k-$NN score with the score computed via the score estimated by domain classifiers as in DARA. We compare DmC with this variant on \emph{medium} source datasets of Ant and Walker under gravity shifts. Table \ref{tab:cls_score} shows that this modification leads to a notable performance drop, suggesting that domain classifiers struggle to accurately estimate domain gaps. These results highlight the effectiveness of $k-$NN estimation in DmC.
\subsubsection{Nearest Neighbor $k$}
We evaluate the impact of $k$ in DmC by testing different values ($k = 1, 5, 10$). As shown in Table \ref{tab:k-nn}, DmC remains robust across varying $k$ values. In our experiments, we set $k = 5$ by default and do not tune it.
\begin{table}[t]
\centering
\caption{Performance comparison with different values of $k$-NN.}
\label{tab:k-nn}
\resizebox{0.75\columnwidth}{!}{%
\begin{tabular}{@{}ll|lll@{}}
\toprule
\multicolumn{1}{l}{\multirow{2}{*}{Source}} & \multicolumn{1}{l}{\multirow{2}{*}{Target}} & \multicolumn{3}{c}{$k$-Nearest Neighbor}                               \\ \cmidrule(l){3-5} 
\multicolumn{1}{c}{}                        & \multicolumn{1}{c}{}                        & \multicolumn{1}{c}{1} & \multicolumn{1}{c}{5} & \multicolumn{1}{c}{10} \\ \midrule
Half-gravity-m                              & me                                           & 48.8±0.7              & \textbf{48.9±0.7}     & 48.5±0.5               \\
Half-gravity-m                              & e                                            & \textbf{49.1±0.6}     & 48.8±1                & 49.0±0.3                 \\
Half-kinematic-m                            & me                                           & \textbf{19.2±5}       & 19.1±1                & 14.6±1.5               \\
Half-kinematic-m                            & e                                            & 13.4±1.2              & 13.1±0.8              & \textbf{13.5±0.9}      \\
Hopp-gravity-m                              & me                                           & 36.9±12               & 40.6±5.5              & \textbf{44.6±10.5}     \\
Hopp-gravity-m                              & e                                            & 40.2±8                & \textbf{51.3±10.6}    & 34.4±9.8               \\
Hopp-kinematic-m                            & me                                           & 73.6±3.1              & \textbf{78.2±5.1}     & 76.6±4.2               \\
Hopp-kinematic-m                            & e                                            & 57.9±11.7             & 59.8±21.8             & \textbf{61.9±17.5}     \\ \bottomrule
\end{tabular}%
}
\end{table}
\section{Conclusion}
In this paper, we tackle the challenge of cross-domain offline RL in scenarios with limited target data, which is widely encountered in many real-world applications. We systematically analyze the limitations of existing methods under this setting, identifying key shortcomings. Building on these insights, we propose DmC, a novel algorithm that leverages $k$-NN estimation to effectively quantify domain gaps and uses this score in a guided diffusion model to generate source samples closer to the target domain. Our approach is compatible with any offline RL method, offering broad applicability. Extensive experiments demonstrate the superior performance of DmC across various benchmarks. A limitation of DmC is that it only adopts a diffusion model for the single-step dynamics model. Extending it to model full trajectory distributions presents an exciting avenue for further exploration and in-depth study.

\bibliography{ecai}

\onecolumn



\section{Missing Theoretical Proofs}\label{appendix:missing_theorem}
In this section, we provide formal proofs omitted from the main paper. 

\subsection{Useful Lemmas}

\begin{lemma}\label{lemma:extended_telescoping}
[Extended telescoping lemma.]
    Let $\mathcal{M}_1 = (\mathcal{S}, \mathcal{A}, P_1, r, \gamma)$ and $\mathcal{M}_2 = (\mathcal{S}, \mathcal{A}, P_2, r, \gamma)$ be two MDPs with different transition dynamics $P_1$ and $P_2$. Given two policies $\pi_1$ and $\pi_2$, we have the following:
    \begin{equation}
        J^{P_1}(\pi_1) - J^{P_2}(\pi_2) = \frac{\gamma}{1 - \gamma}\mathbb{E}_{s, a\sim d^{\pi_1}_{P_1}}\left[\mathbb{E}_{s'\sim P_1, a'\sim \pi_1}\left[Q^{\pi_2}_{P_2}(s', a')\right] - \mathbb{E}_{s'\sim P_2, a'\sim \pi_2}\left[Q^{\pi_2}_{P_2}(s', a')\right]\right].
    \end{equation}
\end{lemma}

This lemma is Lemma C.2 in \cite{xu2023cross} and is the extended version of the telescoping lemma in \cite{luoalgorithmic}. Here, we provide the brief proof.

\emph{Proof.} We define $W_j$ as the expected return when follow the policy $\pi_1$ in  $\mathcal{M}_1$ for the first $j$ steps, then switching to the policy $\pi_1$ and $\mathcal{M}_2$ for the remainder. We have:
\begin{equation}
    W_j:=\sum_{t=0}^{\infty} \gamma^t \mathbb{E}_{\substack{t<j: s_t, a_t \sim P_1, \pi_1 \\t \geq j: s_t, a_t \sim P_2, \pi_2}} \left[ r\left(s_t, a_t\right)\right] = 
    \mathbb{E}_{\substack{t<j: s_t, a_t \sim P_1, \pi_1 \\ t \geq j: s_t, a_t \sim P_2, \pi_2}} \left[ \sum_{t=0}^{\infty} \gamma^t r\left(s_t, a_t\right) \right].
\end{equation}

Note that we have:
\begin{equation}
    \begin{aligned}
        &W_0 = \mathbb{E}_{s, a \sim d^{\pi_2}_{P_2}}\left[r(s_t, a_t)\right] = J^{P_2}(\pi_2), \\
        \text{and } &W_{\infty} = \mathbb{E}_{s, a\sim d^{\pi_1}_{P_1}}\left[r(s_t, a_t)\right] = J^{P_1}(\pi_1).
    \end{aligned}
\end{equation}

We then have the following:
\begin{equation}
    J^{P_1}(\pi_1) - J^{P_2}(\pi_2) = \sum_{j=0}^{\infty}(W_{j+1} - W_j).
\end{equation}

Write $W_j$ and $W_{j+1}$ as the following:
\begin{equation}
    \begin{aligned}
        W_{j} &= R_j + \mathbb{E}_{s_j, a_j \sim P_1, \pi_1}\left[\mathbb{E}_{s_{j+1}, a_{j+1}\sim P_2, \pi_2}\left[\gamma^{j+1}Q^{\pi_2}_{P_2}(s_{j+1}, a_{j+1})\right]\right], \\
        W_{j+1} &= R_j + \mathbb{E}_{s_j, a_j \sim P_1, \pi_1}\left[\mathbb{E}_{s_{j+1}, a_{j+1}\sim P_1, \pi_1}\left[\gamma^{j+1}Q^{\pi_2}_{P_2}(s_{j+1}, a_{j+1})\right]\right]
    \end{aligned}
\end{equation}
Thus, we have:
\begin{equation}
    \begin{aligned}
        J^{P_1}(\pi_1) - J^{P_2}(\pi_2) &= \sum_{j=0}^{\infty}(W_{j+1} - W_j) \\
        &= \sum_{j=0}^{\infty}\gamma^{j+1} \mathbb{E}_{s_j, a_j \sim P_1, \pi_1}\left[\mathbb{E}_{s_{j+1}, a_{j+1}\sim P_1, \pi_1}\left[Q^{\pi_2}_{P_2}(s_{j+1}, a_{j+1})\right] - \mathbb{E}_{s_{j+1}, a_{j+1}\sim P_2, \pi_2}\left[Q^{\pi_2}_{P_2}(s_{j+1}, a_{j+1})\right]\right] \\
        &= \frac{\gamma}{1 - \gamma} \mathbb{E}_{s_j, a_j \sim P_1, \pi_1}\left[\mathbb{E}_{s_{j+1}, a_{j+1}\sim P_1, \pi_1}\left[Q^{\pi_2}_{P_2}(s_{j+1}, a_{j+1})\right] - \mathbb{E}_{s_{j+1}, a_{j+1}\sim P_2, \pi_2}\left[Q^{\pi_2}_{P_2}(s_{j+1}, a_{j+1})\right]\right],
    \end{aligned}
\end{equation}
which concludes the proof.

\begin{lemma}\label{lemma:policy_telescoping}
Let $\mathcal{M} = (\mathcal{S}, \mathcal{A}, P, r, \gamma)$ be the MDP. Given two policies $\pi_1$ and $\pi_2$, we have the following:
\begin{equation}
    J^{P}(\pi_1) - J^P(\pi_2) = \frac{\gamma}{1 - \gamma}\mathbb{E}_{s, a \sim d^{\pi_1, P}, s'\sim P}\left[\mathbb{E}_{a'\sim \pi_1}\left[Q^{\pi_2}_{P}(s', a')\right] - \mathbb{E}_{a'\sim \pi_2}\left[Q^{\pi_2}_{P}(s', a')\right]\right].
\end{equation}
\end{lemma}

This is Lemma B.3 in \cite{lyucross}. We provide the brief proof here.

\emph{Proof.} Using Lemma \ref{lemma:extended_telescoping} and replace $P_1$ and $P_2$ by $P$, we have the following:
\begin{equation}
    \begin{aligned}
        J^{P}(\pi_1) - J^{P}(\pi_2) = \frac{\gamma}{1 - \gamma} \mathbb{E}_{s_j, a_j \sim P, \pi_1}\left[\mathbb{E}_{s_{j+1}, a_{j+1}\sim P, \pi_1}\left[Q^{\pi_2}_{P}(s_{j+1}, a_{j+1})\right] - \mathbb{E}_{s_{j+1}, a_{j+1}\sim P, \pi_2}\left[Q^{\pi_2}_{P}(s_{j+1}, a_{j+1})\right]\right],
    \end{aligned}
\end{equation}
which completes the proof.

\subsection{Proof of Theorems 1}

\begin{theorem}[Performance Bound] Denote $D_{src}$ as the offline source dataset from source domain $\mathcal{M}_{src}$ and $D_{tar}$ as the offline target dataset from target domain $\mathcal{M}_{tar}$. Let the empirical policy in the offline target dataset $D_{tar}$ be $\pi_{D_{tar}} = \frac{\sum_{D_{tar}}\mathds{1}(s, a)}{\sum_{D_{tar}}\mathds{1}(s)}$. Given a policy $\pi$, we have the following:
\begin{equation}
\begin{aligned}
    &J^{P_{tar}}(\pi) - J^{\widehat{P}_{src}}(\pi) \\& \geq - \frac{2r_{\text{max}}\gamma}{(1 - \gamma)^2} \mathbb{E}_{s, a \sim d^{\pi}_{P_{tar}}, s'\sim P_{tar}} \left[ D_{TV}\left(\pi(.|s') || \pi_{D_{tar}}(.|s')\right) \right] -\frac{2r_{\text{max}}\gamma}{(1-\gamma)^2}  \mathbb{E}_{s, a\sim d^{\pi_{D_{tar}}}_{P_{tar}}, s'\sim P_{tar}}\left[D_{TV}\left(\pi(.|s') || \pi_{D_{tar}}(.|s')\right)\right] \\
    &- \frac{2r_{\text{max}}\gamma}{(1-\gamma)^2} \mathbb{E}_{s, a\sim d^{\pi_{D_{tar}}}_{P_{tar}}} \left[\sqrt{1/2D_{KL}(P_{tar}(.|s, a) || \widehat{P}_{src}(.|s, a))}\right],
\end{aligned}
\end{equation}
    
\end{theorem}

\emph{Proof.} We start by converting the performance difference in the following form:
\begin{equation}
    \begin{aligned}
        J^{P_{tar}}(\pi) - J^{\widehat{P}_{src}}(\pi) = \underbrace{\left(J^{P_{tar}}(\pi) - J^{P_{tar}}(\pi_{D_{tar}})\right)}_{\text{(a)}} + \underbrace{\left(J^{P_{tar}}(\pi_{D_{tar}}) - J^{\widehat{P}_{src}}(\pi)\right)}_{(b)}.
    \end{aligned}
\end{equation}

For term $(a)$ in the RHS, based on Lemma \ref{lemma:policy_telescoping}, we have:
\begin{equation}
    \begin{aligned}
        J^{P_{tar}}(\pi) - J^{P_{tar}}(\pi_{D_{tar}}) &=  \frac{\gamma}{1 - \gamma}\mathbb{E}_{s, a \sim d^{\pi}_{P_{tar}}, s'\sim P_{tar}}\left[\mathbb{E}_{a'\sim \pi}\left[Q^{\pi_{D_{tar}}}_{P_{tar}}(s', a')\right] - \mathbb{E}_{a'\sim \pi_{D_{tar}}}\left[Q^{\pi_{D_{tar}}}_{P_{tar}}(s', a')\right]\right] \\
        &\geq - \frac{\gamma}{1 - \gamma} \mathbb{E}_{s, a \sim d^{\pi}_{P_{tar}}, s'\sim P_{tar}}\left|\mathbb{E}_{a'\sim \pi}\left[Q^{\pi_{D_{tar}}}_{P_{tar}}(s', a')\right] - \mathbb{E}_{a'\sim \pi_{D_{tar}}}\left[Q^{\pi_{D_{tar}}}_{P_{tar}}(s', a')\right]\right| \\
        & = - \frac{\gamma}{1 - \gamma} \mathbb{E}_{s, a \sim d^{\pi}_{P_{tar}}, s'\sim P_{tar}}\left|\sum_{a'\in \mathcal{A}}\left(\pi(a'|s') - \pi_{D_{tar}}(a'|s')\right) Q^{\pi_{D_{tar}}}_{P_{tar}}(s', a')\right| \\
        & \geq - \frac{r_{\text{max}}\gamma}{(1 - \gamma)^2} \mathbb{E}_{s, a \sim d^{\pi}_{P_{tar}}, s'\sim P_{tar}}\left|\sum_{a'\in \mathcal{A}}\left(\pi(a'|s') - \pi_{D_{tar}}(a'|s')\right)\right| \\
        &= - \frac{2r_{\text{max}}\gamma}{(1 - \gamma)^2} \mathbb{E}_{s, a \sim d^{\pi}_{P_{tar}}, s'\sim P_{tar}} \left[ D_{TV}\left(\pi(.|s') || \pi_{D_{tar}}(.|s')\right) \right].
    \end{aligned}
\end{equation}

For term $(b)$ in the RHS, based on Lemma \ref{lemma:extended_telescoping}, we have:
\begin{equation}
    \begin{aligned}
        &J^{P_{tar}}(\pi_{D_{tar}}) - J^{\widehat{P}_{src}}(\pi) \\ &= \frac{\gamma}{1 - \gamma}\mathbb{E}_{s, a\sim d^{\pi_{D_{tar}}}_{P_{tar}}}\left[\mathbb{E}_{s'\sim P_{tar}, a'\sim \pi_{D_{tar}}}\left[Q^{\pi}_{\widehat{P}_{src}}(s', a')\right] - \mathbb{E}_{s'\sim \widehat{P}_{src}, a'\sim \pi}\left[Q^{\pi}_{\widehat{P}_{src}}(s', a')\right]\right] \\
        &= \frac{\gamma}{1 - \gamma}\mathbb{E}_{s, a\sim d^{\pi_{D_{tar}}}_{P_{tar}}}\Bigg[\underbrace{\left(\mathbb{E}_{s'\sim P_{tar}, a'\sim \pi_{D_{tar}}}\left[Q^{\pi}_{\widehat{P}_{src}}(s', a')\right] -  \mathbb{E}_{s'\sim P_{tar}, a'\sim \pi}\left[Q^{\pi}_{\widehat{P}_{src}}(s', a')\right]  \right)}_{(c)}  \\
        &+ \underbrace{\left(\mathbb{E}_{s'\sim P_{tar}, a'\sim \pi}\left[Q^{\pi}_{\widehat{P}_{src}}(s', a')\right] - \mathbb{E}_{s'\sim \widehat{P}_{src}, a'\sim \pi}\left[Q^{\pi}_{\widehat{P}_{src}}(s', a')\right] \right)}_{(d)}\Bigg].
    \end{aligned}
\end{equation}

We first bound term $(c)$ as follows:
\begin{equation}
    \begin{aligned}
        &\mathbb{E}_{s'\sim P_{tar}, a'\sim \pi_{D_{tar}}}\left[Q^{\pi}_{\widehat{P}_{src}}(s', a')\right] -  \mathbb{E}_{s'\sim P_{tar}, a'\sim \pi}\left[Q^{\pi}_{\widehat{P}_{src}}(s', a')\right] \\ &= \mathbb{E}_{s'\sim P_{tar}}\left[\sum_{a'\in \mathcal{A}}\left(\pi_{D_{tar}}(a'|s') - \pi(a'|s'))\right)Q^{\pi}_{\widehat{P}_{src}}(s', a')\right] \\
        & \geq - \mathbb{E}_{s'\sim P_{tar}}\left[\sum_{a'\in \mathcal{A}}\left|\pi_{D_{tar}}(a'|s') - \pi(a'|s'))\right|\left|Q^{\pi}_{\widehat{P}_{src}}(s', a')\right|\right] \\
        &\geq -\frac{2r_{\text{max}}}{1 - \gamma} \mathbb{E}_{s'\sim P_{tar}}\left[D_{TV}\left(\pi(.|s') || \pi_{D_{tar}}(.|s')\right)\right].
    \end{aligned}
\end{equation}

For term $(d)$, we have:
\begin{equation}
    \begin{aligned}
        & \mathbb{E}_{s'\sim P_{tar}, a'\sim \pi}\left[Q^{\pi}_{\widehat{P}_{src}}(s', a')\right] - \mathbb{E}_{s'\sim \widehat{P}_{src}, a'\sim \pi}\left[Q^{\pi}_{\widehat{P}_{src}}(s', a')\right] \\ &=
        \mathbb{E}_{a'\sim \pi} \left[\sum_{s'\in \mathcal{S}}\left( P_{tar}(s'|s, a) - \widehat{P}_{src}(s'|s, a)\right)Q^{\pi}_{\widehat{P}_{src}}(s', a')\right] \\
        & \geq -\frac{2r_{\text{max}}}{1-\gamma}  D_{TV}(P_{tar}(.|s, a) || \widehat{P}_{src}(.|s, a)) \\
        & \geq -\frac{2r_{\text{max}}}{1-\gamma} \sqrt{1/2D_{KL}(P_{tar}(.|s, a) || \widehat{P}_{src}(.|s, a))}.
    \end{aligned}
\end{equation}

Thus we can bound term $(b)$ as follows:
\begin{equation}
    \begin{aligned}
        &J^{P_{tar}}(\pi_{D_{tar}}) - J^{\widehat{P}_{src}}(\pi) \\
        &= \frac{\gamma}{1 - \gamma}\mathbb{E}_{s, a\sim d^{\pi_{D_{tar}}}_{P_{tar}}}\Bigg[\underbrace{\left(\mathbb{E}_{s'\sim P_{tar}, a'\sim \pi_{D_{tar}}}\left[Q^{\pi}_{\widehat{P}_{src}}(s', a')\right] -  \mathbb{E}_{s'\sim P_{tar}, a'\sim \pi}\left[Q^{\pi}_{\widehat{P}_{src}}(s', a')\right]  \right)}_{(c)}  \\
        &+ \underbrace{\left(\mathbb{E}_{s'\sim P_{tar}, a'\sim \pi}\left[Q^{\pi}_{\widehat{P}_{src}}(s', a')\right] - \mathbb{E}_{s'\sim \widehat{P}_{src}, a'\sim \pi}\left[Q^{\pi}_{\widehat{P}_{src}}(s', a')\right] \right)}_{(d)}\Bigg] \\
        &\geq -\frac{2r_{\text{max}}\gamma}{(1-\gamma)^2}  \mathbb{E}_{s, a\sim d^{\pi_{D_{tar}}}_{P_{tar}}}\Bigg[\mathbb{E}_{s'\sim P_{tar}}\left[D_{TV}\left(\pi(.|s') || \pi_{D_{tar}}(.|s')\right)\right] + \sqrt{1/2D_{KL}(P_{tar}(.|s, a) || \widehat{P}_{src}(.|s, a))} \Bigg].
    \end{aligned}
\end{equation}

Thus, we have:
\begin{equation}
    \begin{aligned}
        &J^{P_{tar}}(\pi) - J^{\widehat{P}_{src}}(\pi) \\
        &\geq  - \frac{2r_{\text{max}}\gamma}{(1 - \gamma)^2} \mathbb{E}_{s, a \sim d^{\pi}_{P_{tar}}, s'\sim P_{tar}} \left[ D_{TV}\left(\pi(.|s') || \pi_{D_{tar}}(.|s')\right) \right] \\
        &-\frac{2r_{\text{max}}\gamma}{(1-\gamma)^2}  \mathbb{E}_{s, a\sim d^{\pi_{D_{tar}}}_{P_{tar}}}\Bigg[\mathbb{E}_{s'\sim P_{tar}}\left[D_{TV}\left(\pi(.|s') || \pi_{D_{tar}}(.|s')\right)\right] + \sqrt{1/2D_{KL}(P_{tar}(.|s, a) || \widehat{P}_{src}(.|s, a))} \Bigg]  \\
        &= - \frac{2r_{\text{max}}\gamma}{(1 - \gamma)^2} \mathbb{E}_{s, a \sim d^{\pi}_{P_{tar}}, s'\sim P_{tar}} \left[ D_{TV}\left(\pi(.|s') || \pi_{D_{tar}}(.|s')\right) \right] -\frac{2r_{\text{max}}\gamma}{(1-\gamma)^2}  \mathbb{E}_{s, a\sim d^{\pi_{D_{tar}}}_{P_{tar}}, s'\sim P_{tar}}\left[D_{TV}\left(\pi(.|s') || \pi_{D_{tar}}(.|s')\right)\right] \\
        &- \frac{2r_{\text{max}}\gamma}{(1-\gamma)^2} \mathbb{E}_{s, a\sim d^{\pi_{D_{tar}}}_{P_{tar}}} \left[\sqrt{1/2D_{KL}(P_{tar}(.|s, a) || \widehat{P}_{src}(.|s, a))}\right],
    \end{aligned}
\end{equation}
 which completes the proof.

\section{Cross-Domain Offline RL with Limited Target Samples}\label{appendix:cross_domain_discussion}
In this section, we provide detailed discussions about the challenge of the Cross-domain Offline RL with limited target data setting.
\subsection{Datasets Imbalanced Problem}
\begin{figure*}[!htbp]
        \begin{subfigure}[b]{0.32\textwidth}
                \includegraphics[width=1\textwidth]{images/sas_tar_probs.png}
        \end{subfigure}%
        \begin{subfigure}[b]{0.32\textwidth}
                \includegraphics[width=1\textwidth]{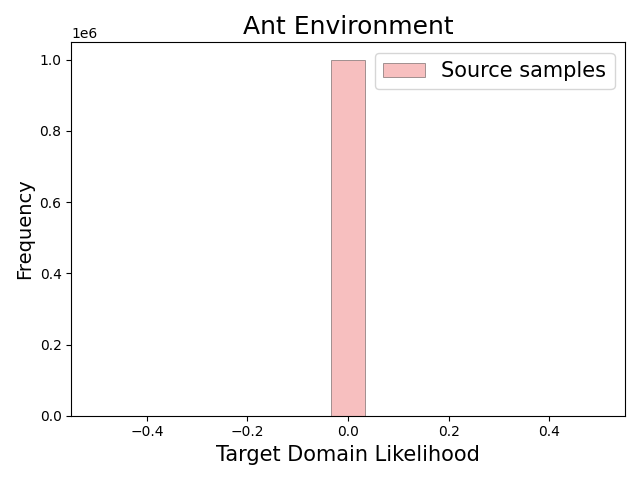}
        \end{subfigure}%
        \begin{subfigure}[b]{0.32\textwidth}
                \includegraphics[width=1\textwidth]{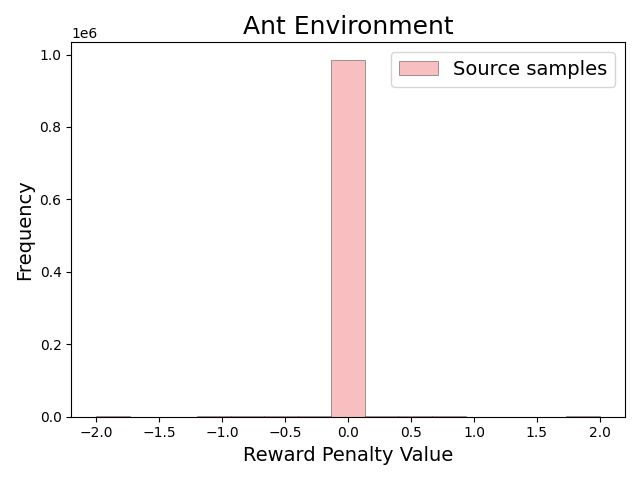}
        \end{subfigure}%
        \caption{(Left) Histogram of the predicted target likelihood of the domain classifiers in DARA for the source sample in the source dataset. (Middle) The histogram of the predicted target likelihood of the CVAE dynamics target model proposed in BOSA. (Right) The histogram of the reward penalty values computed using domain classifiers in DARA.}
\end{figure*}
\bigskip
In cross-domain offline RL, a critical challenge is effectively quantifying the domain gap between the source and target domains. Accurate measurement of this gap enables subsequent strategies to leverage source samples to enhance policy learning for the target domain. Prior works have proposed various approaches for this purpose, including training domain classifiers \cite{liu2022dara}, or estimating target models \cite{liu2024beyond}. However, these methods often rely on training parametric models such as neural networks, which are prone to overfitting in scenarios with limited target data and significant dataset imbalances. For instance, limited target samples can cause classifiers to develop biases, favoring data that appears more frequently while failing to capture the true dynamics gap.
To investigate potential overfitting issues, we train the domain classifiers from DARA \cite{liu2022dara} and the CVAE dynamics model from BOSA \cite{liu2024beyond} in Ant environment with \emph{medium} source and \emph{expert} target datasets under the gravity shift. We then evaluate the predicted target likelihood of these models on samples from the source dataset $D_{src}$. Figures 1-Left and 1-Middle show the histograms of the predicted target likelihoods obtained from the learned domain classifiers and the dynamics model. Both exhibit significant overfitting, as they predominantly output uniform predictions across all source samples. Additionally, the reward penalties computed via DARA's domain classifiers are near zero, providing little to no useful signal for policy learning, as shown in Figure 1-Right. In addition to the overfitting issues, the imbalance between source and target data can bias the learned policy, potentially causing it to overfit the source data during training.
\subsection{Partial Overlapping between Domains}
Leveraging source samples is critical for enhancing policy learning and improving sample efficiency in cross-domain offline RL. However, discrepancies in dynamics between the source and target domains, along with potential policy shifts during data collection processes, not all source data is close to the target data. To investigate how the source data is close to the target data, we conduct experiments in the Ant environment with \emph{medium} source and \emph{expert} target datasets under gravity shift. Specifically, we compute the distances from each source sample to its nearest neighbor in the target dataset $D_{tar}$ and the distances from each target sample to its nearest neighbor within $D_{tar}$. The histograms in Figure \ref{fig:log_distance} reveal only partial overlap between two histograms, highlighting that only a subset of the source dataset is beneficial for target policy learning. Previous methods \cite{liu2024beyond, wencontrastive} addressed this by filtering out source samples far from the target domain. However, this approach reduces the number of usable source samples, ultimately impacting the sample efficiency of the algorithms. This raises an important question: ``Can we further improve sample efficiency in cross-domain offline RL by generating additional source data that closely aligns with the target domain?"
\begin{figure}[h]
    \centering
    \includegraphics[width=0.5\linewidth]{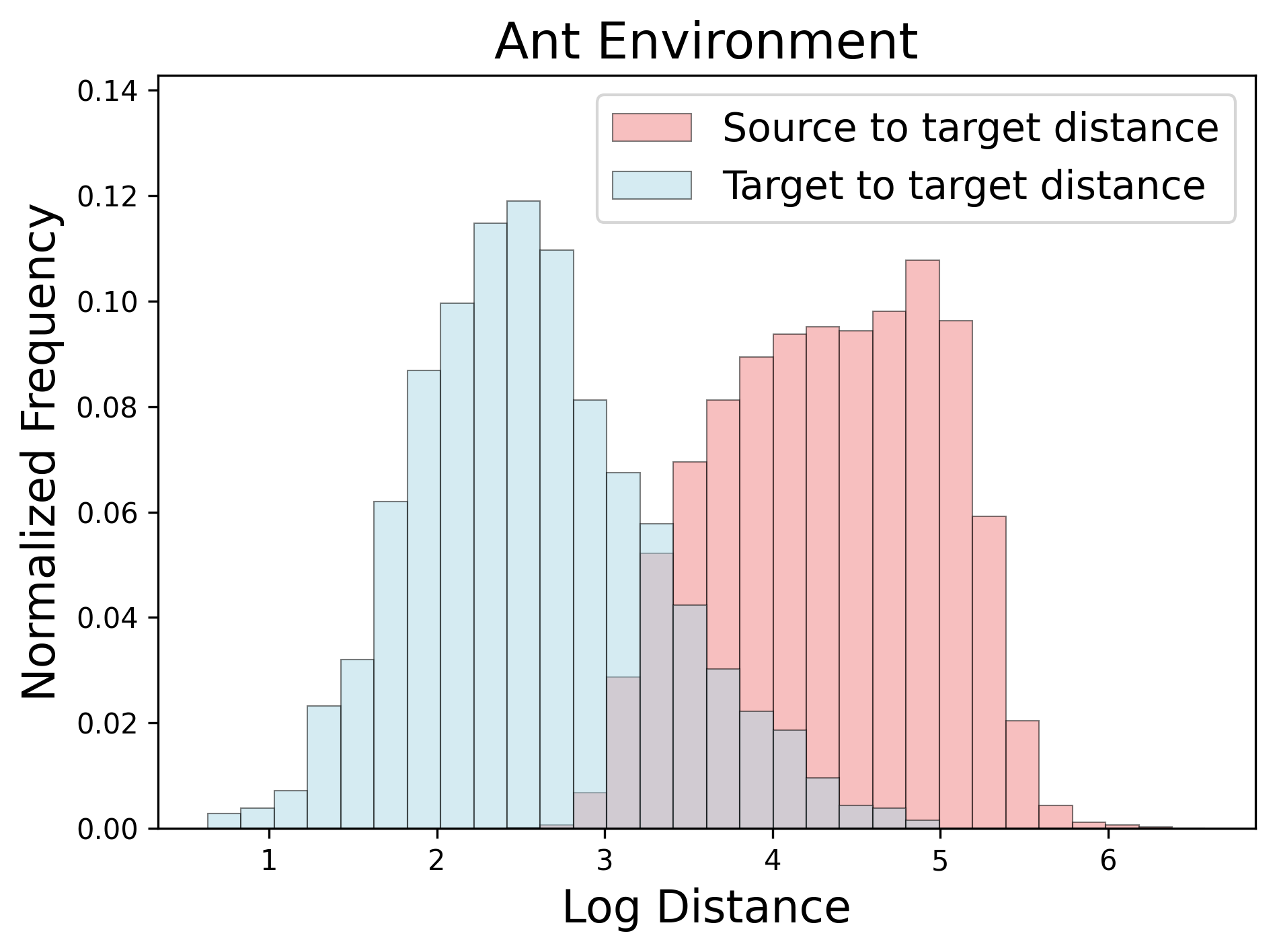}
    \caption{Nearest Neighbor Distance histograms. Blue shows the distance of source samples to their nearest target samples. Pink shows the distance of target samples to their nearest target samples.}
    \label{fig:log_distance}
\end{figure}

\section{Environment Setting}\label{appendix:environment}
This section provides a detailed description of the environment settings we use in our experiments. 

\subsection{Source Environments}

We conduct our experiments on four Mujoco environments: Ant, Halfcheetah, Hopper, and Walker from the Openai-Gym libraries \cite{todorov2012mujoco,brockman2016openai}. For our offline source, we use datasets from the D4RL benchmark \cite{fu2020d4rl}, which provides three types of datasets for each environment: medium, medium-replay, and medium-expert. Specifically, the medium datasets consist of experiences collected from an SAC policy that was early-stopped after 1 million steps. The medium-replay datasets contain the replay buffer of a policy trained to achieve the medium agent's performance. The medium-expert datasets are constructed by combining medium and expert data in a 50-50 ratio. Notably, the sizes of these datasets vary significantly; for example, the medium datasets contain 1 million samples, while the medium-replay datasets may contain as few as 100,000 samples.

\subsection{Target Environments}
In this paper, we consider the gravity shift and kinematic shift as the dynamics shift between the source domain and the target domain. Gravity shift refers to differences in gravitational forces acting on the source and target robots. To simulate this shift, we modify the gravitational acceleration parameter in the environment while keeping its direction consistent with the default configuration in MuJoCo. Specifically, we retain the default gravitational acceleration value for the source domain and set the value in the target domain to be half of that in the source domain. On the other hand, when referring to kinematic shift, we are referring to altering some joints of the simulated robots. 
We provide the XML modifications in Section \ref{appendix:sec_gravity_shift} and Section \ref{appendix:section_kinematic_shift}, along with the XML files of the agents in our supplementary materials.

The target datasets are collected following a procedure similar to D4RL. We consider three dataset quality levels: \emph{expert}, \emph{medium}, and \emph{medium-expert}. Each dataset consists of five trajectories, totaling 5,000 transitions, to reflect the limited target data setting. The \emph{expert} datasets are collected using the last checkpoint of the trained policy, while the \emph{medium} datasets use a policy performing at approximately 1/2 or 1/3 of the expert policy's performance. The \emph{medium-expert} datasets are constructed by combining two trajectories from the \emph{medium} dataset and three from the \emph{expert} dataset.
Finally, we use the target datasets collected by \cite{lyu2025crossdomain} in our experiments as we consider the same target domain setting as them.


\subsubsection{Gravity shift}\label{appendix:sec_gravity_shift}
The XML files of Ant, Halfcheetah, Hopper, and Walker are modified as follows:
    \begin{lstlisting}[language=XML]
    <option gravity="0 0 -4.905" timestep="0.01"/>
    \end{lstlisting}
    
\subsubsection{Kinematic shift} \label{appendix:section_kinematic_shift}
The kinematic shifts of the robot occurred at different parts, detailed below:
\begin{itemize}
    \item \textbf{Ant-kinematic}: The rotation angles of the joints on the hips of two legs in the ant robot are adjusted from \([-30, 30]\) to \([-0.3, 0.3]\).
    \begin{lstlisting}[language=XML]
    # hip joints of the front legs
    <joint axis="0 0 1" name="hip_1" pos="0.0 0.0 0.0" range="-0.30 0.30" type="hinge"/>
    <joint axis="0 0 1" name="hip_2" pos="0.0 0.0 0.0" range="-0.30 0.30" type="hinge"/>
    \end{lstlisting}
    \item \textbf{HalfCheetah-kinematic}: The rotation angle of the joint on the thigh of the robot’s back leg is adjusted from \([-0.52, 1.05]\) to \([-0.0052, 0.0105]\).
    \begin{lstlisting}[language=XML]
    # back thigh
    <joint axis="0 1 0" damping="6" name="bthigh" pos="0 0 0" range="-.0052 .0105" stiffness="240" type="hinge"/>
    \end{lstlisting}
    \item \textbf{Hopper-kinematic}: The rotation angle of the head joint is adjusted from \([150, 0]\) to \([0.15, 0]\) and the rotation angle of the joints on the robot's foot is modified from \([-45, 45]\) to \([-18, 18]\).
    \begin{lstlisting}[language=XML]
    # head joint
    <joint axis="0 -1 0" name="thigh_joint" pos="0 0 1.05" range="-0.150 0" type="hinge"/>
    # foot joint
    <joint axis="0 -1 0" name="foot_joint" pos="0 0 0.1" range="-18 18" type="hinge"/>
    \end{lstlisting}
    \item \textbf{Walker-Kinematic}: The rotation angle of the foot joint on the robot’s right leg is modified from \([-45, 45]\) to \([-0.45, 0.45]\).
    \begin{lstlisting}[language=XML]
    # right foot
    <joint axis="0 -1 0" name="foot_joint" pos="0 0 0.1" range="-0.45 0.45" type="hinge"/>
    \end{lstlisting}
\end{itemize}


\subsection{Evaluation Metric}
To evaluate the adaptation performance of the learned policy in the target domain, we use the \emph{normalized score} (NS) metric as similar in \cite{fu2020d4rl, lyu2024odrlabenchmark}. The normalized score of a given policy in the target domain is computed as follows:
\begin{equation}
    NS = \frac{J - J_r}{J_e - J_r} \times 100,
\end{equation}
where $J, J_e, J_r$ denotes the return of the given, expert, and random policies in the target domain. We list the reference scores in Table \ref{tab:ref_scores} and provide the minimum return, the maximum return, and the average return of the trajectories in each target dataset in Table \ref{tab:returns_of_the_dataset}.

\begin{table}[ht]
\centering
\caption{The reference scores of the Mujoco datasets under the gravity shifts. $J_r$ denotes the performance of the random policy in the target domain, and $J_e$ denotes the performance of the expert policy in the target domain. The reference scores are used to compute the \emph{normalized score} for evaluating the performance of the learned policies.}
\label{tab:ref_scores}
\begin{tabular}{@{}lccc@{}}
\toprule
Environment & Dynamic Shift Type & Reference score $J_r$ & Reference score $J_e$ \\ \midrule
Ant         & Gravity            & -325.6                & 4317.065              \\
Halfcheetah & Gravity            & -280.18               & 9509.15               \\
Hopper      & Gravity            & -26.336               & 3234.3                \\
Walker      & Gravity            & 10.079                & 5194.713              \\
Ant         & Kinematic          & -325.6                & 5122.57               \\
Halfcheetah & Kinematic          & -280.18               & 7065.03               \\
Hopper      & Kinematic          & -26.336               & 2842.73               \\
Walker      & Kinematic          & 10.079                & 3257.51               \\ \bottomrule
\end{tabular}
\end{table}

\begin{table}[ht]
\centering
\caption{Trajectories return information of the target datasets. We report the trajectory return statistics of the target domain datasets, including the minimum return (min return), maximum return (max return), and average return.}
\label{tab:returns_of_the_dataset}
\begin{tabular}{@{}lllrrr@{}}
\toprule
Task Name   & Dynamics shift & Dataset type  & Min return & Max return & Average return \\ \midrule
Ant         & Gravity        & medium        & 377.10     & 3247.66    & 2314.45        \\
Ant         & Gravity        & medium-expert & 377.10     & 4511.55    & 2131.79        \\
Ant         & Gravity        & expert        & 335.28     & 4584.53    & 3365.35        \\
Ant         & Kinematic      & medium        & 2826.00    & 3111.98    & 3017.82        \\
Ant         & Kinematic      & medium-expert & 2826.00    & 5122.58    & 4240.99        \\
Ant         & Kinematic      & expert        & 5009.82    & 5122.57    & 5072.50        \\
Halfcheetah & Gravity        & medium        & 4179.82    & 4383.32    & 4296.27        \\
Halfcheetah & Gravity        & medium-expert & 4342.78    & 8243.03    & 6567.94        \\
Halfcheetah & Gravity        & expert        & 7846.18    & 8339.18    & 8131.54        \\
Halfcheetah & Kinematic      & medium        & 2709.52    & 2782.61    & 2755.50        \\
Halfcheetah & Kinematic      & medium-expert & 2709.52    & 7065.04    & 5298.61        \\
Halfcheetah & Kinematic      & expert        & 6951.27    & 7065.04    & 6998.93        \\
Hopper      & Gravity        & medium        & 1784.88    & 2885.13    & 2367.66        \\
Hopper      & Gravity        & medium-expert & 2416.82    & 4143.63    & 3297.79        \\
Hopper      & Gravity        & expert        & 3745.59    & 4186.19    & 4051.07        \\
Hopper      & Kinematic      & medium        & 1849.06    & 1886.89    & 1870.16        \\
Hopper      & Kinematic      & medium-expert & 1868.20    & 2842.17    & 2452.67        \\
Hopper      & Kinematic      & expert        & 2840.97    & 2842.73    & 2841.83        \\
Walker      & Gravity        & medium        & 2421.98    & 3444.63    & 2897.85        \\
Walker      & Gravity        & medium-expert & 3144.32    & 5166.62    & 4415.11        \\
Walker      & Gravity        & expert        & 5159.51    & 5219.14    & 5174.51        \\
Walker      & Kinematic      & medium        & 1415.69    & 2223.17    & 2026.49        \\
Walker      & Kinematic      & medium-expert & 1415.69    & 3257.51    & 2442.82        \\
Walker      & Kinematic      & expert        & 2874.92    & 3257.51    & 3077.19        \\ \bottomrule
\end{tabular}
\end{table}

\section{Algorithm’s Implementations}\label{appendix:algo_implement}
In this section, we provide a details implementation of our method, DmC, and the baselines we use in the experiments. Specifically, we use the report  We report the hyperparameter in Table \ref{tab:hyper}. 
Our implementation is based on ODRL \cite{lyu2024odrlabenchmark}, Synther \cite{lu2023synthetic}, and CleanDiffuser \cite{cleandiffuser}.

\subsection{DmC} In our implementation, we use the FAISS library \cite{douze2024faiss, johnson2019billion} to compute the $k$-NN estimation.  We compute the $k$-NN based score for all source samples in the source dataset $D_{src}$ at the beginning. The computation time is only within 1 minute for 1 million source samples and 5 thousand target samples. 

After computing the score for each source sample, we train our diffusion model using the offline source dataset with the corresponding scores as the conditional context $y$. We use the design of EDM \cite{karras2022elucidating} for our diffusion model, it has demonstrated its performance in previous offline RL works \cite{lu2023synthetic}. Specifically, we employ a fixed noise schedule $\sigma_{\text{max}}=\sigma^{T}>\dots>\sigma^{1}>\sigma^{0}=0$ and diffuse clean data samples using the transition $q(x^t|x_{src, i})=\mathcal{N}\left(x_{src, i}, (\sigma^t)^2\mathrm{I}\right)$. Then, we train a denoiser $G_\theta(\cdot)$ to directly predict the clean samples using classifier-free training \cite{ho2022classifier}:
\begin{equation}
   \min_\theta \mathbb{E}\left[\left\Vert x_{src, i}-G_{\theta}\left(x_{src, i}+\sigma^{t}\epsilon,m\cdot\rho_k(x_{src, i}),\sigma^{t}\right)\right\Vert ^{2}\right]
\end{equation}
where $x_{src, i}\sim D_{src}$, $\epsilon\sim\mathcal{N}(0,\mathrm{I})$, and $t\sim\mathcal{U}[0,T]$. The score $\rho_k(x_{src, i})$ is randomly masked during training via the null token $m$, which is sampled from the Bernoulli distribution with $p\left(m=\emptyset\right)=0.25$. We then used the trained diffusion model to generate source samples that are close to the target domain by setting the value of the conditional context $y$ to be higher than the top $\kappa\%$ of the source sample scores in the source dataset $D_{src}$. Specifically, we uniformly sample a value $\chi$ from the range $[\kappa, 100]$ and set $y$ to the $\chi$-th quantile of the source dataset’s score distribution. The number of generated samples is 1 million and fixed for all environments. Finally, the generated samples are combined with the source dataset for later policy learning. We use the IQL \cite{kostrikovoffline} as the backbone for DmC. 

We summarize our proposed method in Algorithm 1 in the main paper. Based on the analysis in Section 5.4 in the main paper, to reduce the dynamics gaps and tighten the bound in Theorem 1, we select the source samples that have the highest scores during training to update the Q value function. Then, we further use the score as the weighting value for the source data during the training of the Q value function. This approach ensures the policy adaptively emphasizes source samples that closely align with the target domain, improving adaptation performance.

As we mentioned, the imbalanced dataset could bias the policy to the source samples. Thus, we employ additional policy regularization to ensure the learned policy is close to the support areas of the target dataset. Similar to \cite{wu2022supported}, we learn a conditional VAE, denoted as $\hat{\pi}_{tar}^b(a|s)$,  to model the behavior target policy $\pi_{tar}^{b}(a|s)$. Thus, we optimize the policy with the loss function as follows:
\begin{equation}
\label{eq:appendix_pi}
    \mathcal{L}_{\pi}^{\text{reg}} = \mathcal{L}_{\pi} 
- \lambda \mathbb{E}_{s \sim D_{src}\bigcup D_{tar}}\left[\log \hat{\pi}_{tar}^b\left(\pi(.|s)|s\right) \right],
\end{equation}
where $\mathcal{L}_\pi$ is the policy loss of the offline RL method, $\lambda$ is the coefficient controlling the strength of the additional policy regularization. In our implementation, we use the default config for CVAE as presented in \citet{wu2022supported}
.
\subsection{DARA} DARA \cite{liu2022dara} use the learned domain classifiers to compute the reward penalty $\Delta_r$ and correct the source samples using the following:
\begin{equation}
    \hat{r} = r - \alpha \Delta_r,
\end{equation}
where $\alpha$ is the penalty coefficient that controls the strength of the reward penalty term. The reward penalty is clipped to $[-10, 10]$ following the original paper \cite{liu2022dara}. We report the results of DARA with IQL as the backbone s the fair comparison.

\subsection{BOSA} BOSA \cite{liu2024beyond} handles the dynamics shift problems using a supported policy and value optimization. The value function (critic) in BOSA is learned with the following objective:
\begin{equation}
\begin{aligned}
\min _{Q_\phi} \mathcal{L}_{\text {mix }}\left(Q_\phi\right):= & \mathbb{E}_{\left(\mathbf{s}, \mathbf{a}, r, \mathbf{s}^{\prime}\right) \sim \mathcal{D}_{\text {mix }}, \mathbf{a}^{\prime} \sim \pi_\theta\left(\mathbf{a}^{\prime} \mid \mathbf{s}^{\prime}\right)}\left[\delta\left(Q_\phi\right) \cdot \mathds{1}\left(\hat{P}_{\text {target }}\left(\mathbf{s}^{\prime} \mid \mathbf{s}, \mathbf{a}\right)>\epsilon_{\mathrm{th}}^{\prime}\right)\right]  +\mathbb{E}_{(\mathbf{s}, \mathbf{a}) \sim \mathcal{D}_{\text {source }}}\left[Q_\phi(\mathbf{s}, \mathbf{a})\right],
\end{aligned}
\end{equation}
where $\delta\left(Q_\phi\right) = \left(Q_{\phi}(s, a) - r - Q_{\bar{\phi}}(s, a)\right)^2$, $\mathds{1}$ is the indicator function, $\hat{P}_{tar}$ is the learned target transition dynamics, $\epsilon_{\text{th}}$ is the selection threshold. CVAE is used to model the target dynamics model. 

\subsection{IGDF} IGDF \cite{wencontrastive} measure the dynamics mismatch between two domains via the mutual information estimated via contrastive learning. IGDF uses a score function $h$, using the target dataset as the positive sample and the source dataset as the negative sample. Then, based on the score function, IGDF updates the value function using the following objective:
\begin{equation}
\mathcal{L}_{Q}=\frac{1}{2} \mathbb{E}_{D_{\mathrm{tar}}}\left[\left(Q_\theta-\mathcal{T} Q_\theta\right)^2\right]+\frac{1}{2} \alpha \cdot h\left(s, a, s^{\prime}\right) \mathbb{E}_{\left(s, a, s^{\prime}\right) \sim D_{\mathrm{src}}}\left[\mathds{1}\left(h\left(s, a, s^{\prime}\right)>h_{\xi \%}\right)\left(Q_\theta-\mathcal{T} Q_\theta\right)^2\right],
\end{equation}
where $\alpha$ is the weighting coefficient for the TD in the source data, and $\xi$ is the data selection ratio.

\begin{table}[ht!]
\centering
\caption{Hyperparameter setup for DmC and the baselines.}
\label{tab:hyper}
\begin{tabular}{@{}llllllllllr@{}}
\toprule
                                  & \textbf{Hyperparams}                              &  &  &  &  &  &  &  &  & \textbf{Value} \\ \midrule
\multirow{11}{*}{\textbf{Shared}} & Actor network                                     &  &  &  &  &  &  &  &  & (256,256)      \\
                                  & Critic network                                    &  &  &  &  &  &  &  &  & (256,256)      \\
                                  & Learning rate                                     &  &  &  &  &  &  &  &  & 3e-4           \\
                                  & Discounted factor                                 &  &  &  &  &  &  &  &  & 0.99           \\
                                  & Buffer size                                       &  &  &  &  &  &  &  &  & 1e6            \\
                                  & Activation                                        &  &  &  &  &  &  &  &  & ReLU           \\
                                  & Target update coefficient                         &  &  &  &  &  &  &  &  & 5e-3           \\
                                  & Batch size for source and target                  &  &  &  &  &  &  &  &  & 128            \\
                                  & Temperature coefficient                           &  &  &  &  &  &  &  &  & 0.2            \\
                                  & Max log std                                       &  &  &  &  &  &  &  &  & 2              \\
                                  & Min log std                                       &  &  &  &  &  &  &  &  & -20            \\ \midrule
\multirow{2}{*}{\textbf{DARA}}    & Domain classifiers network                        &  &  &  &  &  &  &  &  & (256,256)      \\
                                  & Reward penalty coefficient $\alpha$               &  &  &  &  &  &  &  &  & 0.1            \\ \midrule
\multirow{4}{*}{\textbf{BOSA}}    & Policy regularization coefficient $\lambda_{\pi}$ &  &  &  &  &  &  &  &  & 0.1            \\
                                  & Transition coefficient $\lambda_{transition}$     &  &  &  &  &  &  &  &  & 0.1            \\
                                  & Threshold parameter $\epsilon, \epsilon'$         &  &  &  &  &  &  &  &  & log(0.01)      \\
                                  & Value weight                                      &  &  &  &  &  &  &  &  & 0.1            \\ \midrule
\multirow{5}{*}{\textbf{IGDF}}    & Representation dimension                          &  &  &  &  &  &  &  &  & \{16,64\}      \\
                                  & Contrastive encoder network                       &  &  &  &  &  &  &  &  & (256,256)      \\
                                  & Encoder training steps                            &  &  &  &  &  &  &  &  & 7000           \\
                                  & Importance coefficient                            &  &  &  &  &  &  &  &  & 1.0            \\
                                  & Data selection ratio $\xi\%$                      &  &  &  &  &  &  &  &  & 75\%           \\ \midrule
\textbf{SRPO}                              & Discriminator network                             &  &  &  &  &  &  &  &  & (256, 256)     \\
                                  & Data selection ratio                              &  &  &  &  &  &  &  &  & 0.5            \\
                                  & Reward coefficient                                &  &  &  &  &  &  &  &  & \{0.1, 0.3\}   \\ \midrule
\textbf{OTDF}                              & CVAE training steps                               &  &  &  &  &  &  &  &  & 10000          \\
                                  & CVAE learning rate                                &  &  &  &  &  &  &  &  & 1e-3           \\
                                  & Cost function                                     &  &  &  &  &  &  &  &  & cosin          \\
                                  & Data filtering ratio                              &  &  &  &  &  &  &  &  & 80\%           \\
                                  & Policy coefficient                                &  &  &  &  &  &  &  &  & \{0.1, 0.5\}   \\ \midrule
\textbf{DmC}                               & $k$th nearest neighbor                            &  &  &  &  &  &  &  &  & 5              \\
                                  & Data selection ratio                             &  &  &  &  &  &  &  &  & 50\%           \\
                                  & Policy regularization coefficient                 &  &  &  &  &  &  &  &  & 0.1            \\
                                  & Score threshold for guided sampling               &  &  &  &  &  &  &  &  & 90\%           \\ \bottomrule
\end{tabular}

\end{table}

\subsection{IQL} IQL \cite{kostrikovoffline} is a state-of-the-art offline RL algorithm. It updates the state-value function and state-action value function via the following objectives:

\begin{equation}
\begin{aligned}
&\mathcal{L}_V=\mathbb{E}_{(s, a) \sim \mathcal{D}_{\mathrm{src}} \cup \mathcal{D}_{\mathrm{tar}}}\left[L_2^\tau\left(Q_{\theta^{\prime}}(s, a)-V_\psi(s)\right)\right], \\
&\mathcal{L}_Q=\mathbb{E}_{\left(s, a, r, s^{\prime}\right) \sim \mathcal{D}_{\mathrm{src}} \cup \mathcal{D}_{\mathrm{tar}}}\left[\left(r(s, a)+\gamma V_\psi\left(s^{\prime}\right)-Q_\theta(s, a)\right)^2\right],
\end{aligned}
\end{equation}
where $L_{2}^\tau(u) = |\tau - \mathds{1}(u<0)| u^2$.
The policy is learned using the advantage-weighted behavior cloning objective:
\begin{equation}
    \mathcal{L}_{\pi} = \mathbb{E}_{D_{src} \bigcup D_{tar}}\left[\exp\left(\beta \times A(s, a)\right)\log \pi(a|s)\right],
\end{equation}
where $A(s, a) = Q(s, a) - V(s, a)$, and $\beta$ is the inverse temperature coefficient.

\subsection{SRPO}
SRPO \cite{xue2024state} proposes optimizing the policy by solving the following constrained optimization problem:

\begin{equation}
\max_\pi \mathbb{E}_{s_t, a_t \sim \tau_\pi} \left[ \sum_{t=0}^\infty \gamma^t r(s_t, a_t) \right] \quad \text{s.t.} \quad D_{KL}(d_\pi(\cdot) \parallel \zeta(\cdot)) < \epsilon, 
\end{equation}

where \(\tau_\pi\) is the trajectory induced by policy \(\pi\), \(d_\pi(\cdot)\) is the stationary state distribution of policy \(\pi\), and \(\zeta(\cdot)\) represents the optimal state distribution under different environment dynamics.

This problem can be reformulated into an unconstrained optimization problem using Lagrange multipliers, where the logarithm of the probability density ratio, \(\lambda \log \frac{\zeta(s_t)}{d_\pi(s_t)}\), is added to the standard reward function. Based on this formulation, SRPO samples a batch of size \(N\) from two offline datasets, \(D_\text{src}\) and \(D_\text{tar}\), and ranks the transitions based on state values. A proportion \(\rho N\) (in the paper \(\rho = 0.5\) is used)  of samples with high state values is labeled as real data, while the remaining samples are labeled as fake data.

A discriminator \(D_\delta(\cdot)\) is then trained to distinguish between these samples, and the rewards are modified as follows:

\begin{equation}
\hat{r}_\text{SRPO} = r + \lambda \cdot \frac{D_\delta(s)}{1 - D_\delta(s)}, 
\end{equation}
where \(\lambda\) is the reward coefficient.

\subsection{OTDF}
OTDF \cite{lyu2025crossdomain} attempts to solve the problem of estimating domain gaps in the limited target data setting with the optimal transport. Specifically, they use cousin distance as the cost function for solving optimal transport problems between the datasets of two domains. After obtaining the optimal coupling $\mu*$, they measure the deviation of the source domain to the target domain as follows:
\begin{equation}
    d(u_t) = -\sum^{|D_{tar}|}_{t'=1}C(u_t, u_{t'})\mu*_{t, t'}, u_t = (s^t_{src}, a^t_{src}, (s'_{src})^t) \sim D_{src},
\end{equation}
where $C$ is the cost function.
They then use $d(u_t)$ to select good source data for policy training via the source data filtering. Additionally, they also employ a policy regularization to ensure the learned policy is close to the support of the target dataset. 

\subsection{Hyperparamters}
We adopt the hyperparameters reported in \citet{lyu2025crossdomain} for baseline methods. For DmC, we set $k=5$ for $k$-NN estimation, use a 50\% data selection ratio, a policy regularization coefficient of 0.1, and a 90\% score threshold for guided sampling from the diffusion model. We note that without bothering the hyperparameter tuning, DmC achieves strong performance across diverse tasks with a single set of hyperparameters.

\subsection{Computing Infrastructure}
We use Python 3.9, Gym 0.23.1, Mujoco 2.3.2 and D4RL 1.1. All experiments are conducted on a Ubuntu 22.04 server with CUDA version 12.2. We report the computing infrastructure that we use to run our experiments in Table \ref{tab:server}.

\begin{table}[H]
\centering
\begin{tabular}{@{}ccccc@{}}
\toprule
\textbf{CPU}                   & \textbf{Number of CPU Cores} & \textbf{GPU} & \textbf{VRAM} & \textbf{RAM} \\ \midrule
Intel(R) Xeon(R) Gold 6248 CPU & 10                           & V100         & 32 GB         & 377 GB       \\ \bottomrule
\end{tabular}
\caption{Computing infrastructure.}
\label{tab:server}
\end{table}

\section{More Experimental Results}\label{appendix:more_exps}
In this section, we present additional experimental results omitted from the main text due to space constraints. We use the same number reported in \citet{lyu2025crossdomain} since we consider similar gravity and kinematic shift settings. We present the performance comparison between DmC and the other baselines under kinematic shift and report additional results on the impact of the guided-diffusion model. Furthermore, we conduct ablation studies on DmC's hyperparameters to provide deeper insights into its behavior and effectiveness.
\subsection{Missing Results on Kinematic Shift Tasks}
We present the performance comparison between DmC and other methods under kinematic shifts in Table \ref{tab:main_kinematic}. We observe DmC excels in \textbf{29} out of 36 tasks, surpassing IQL performance by \textbf{59.4\%}, and achieves a total normalized score of \textbf{1902.2}, compared to 1547.6 of the second best method OTDF. Besides that, we observe the cross-domain RL methods that involve training neural networks to estimate the domain gaps fail to bring performance improvement compared to IQL. These results validate the effectiveness of DmC in cross-domain offline RL with limited target data.

\begin{table*}[ht]
\centering
\caption{Results in kinematic shift tasks. We report normalized scores and their standard deviations in the target domain, averaged over five random seeds. The best score is bold. Half=Halfcheetah, Hopp=Hopper, m=medium, me=medium-expert, mr=medium-replay.}
\label{tab:main_kinematic}
\begin{tabular}{@{}ll|lllllll@{}}
\toprule
\multicolumn{1}{c}{Source} & \multicolumn{1}{c|}{Target} & \multicolumn{1}{c}{IQL} & \multicolumn{1}{c}{DARA} & \multicolumn{1}{c}{BOSA} & \multicolumn{1}{c}{SRPO} & \multicolumn{1}{c}{IGDF} & \multicolumn{1}{c}{OTDF} & \multicolumn{1}{c}{DmC} \\ \midrule
Ant-m                      & medium                      & 50.0±5.6                & 42.3±7.6                 & 20.9±2.6                 & 50.5±6.7                 & 54.5±1.3                 & 55.4±0.0                 & \textbf{62.1±0.6}       \\
Ant-m                      & medium-expert               & 57.8±7.2                & 54.1±3.8                 & 31.7±7.0                 & 54.9±1.3                 & 54.5±4.6                 & 60.7±3.6                 & \textbf{68.9±1.0}       \\
Ant-m                      & expert                      & 59.6±18.5               & 54.2±11.3                & 45.4±8.6                 & 45.5±9.3                 & 49.4±14.6                & 90.4±4.8                 & \textbf{92.1±3.5}       \\
Ant-me                     & medium                      & 49.5±4.1                & 44.7±4.3                 & 19.0±8.0                 & 41.3±8.1                 & 41.8±8.8                 & 50.2±4.3                 & \textbf{60.6±1.3}       \\
Ant-me                     & medium-expert               & 37.2±2.0                & 33.3±7.0                 & 6.4±2.5                  & 32.8±8.0                 & 41.5±4.9                 & 48.8±2.7                 & \textbf{60.4±3.7}       \\
Ant-me                     & expert                      & 18.7±8.1                & 17.8±23.6                & 14.5±9.0                 & 35.2±15.5                & 14.4±22.9                & \textbf{78.4±12.2}       & 76.0±4.1                \\
Ant-mr                     & medium                      & 43.7±4.6                & 42.0±5.4                 & 19.0±1.8                 & 45.3±5.1                 & 41.4±5.0                 & 52.8±4.4                 & \textbf{61.9±0.5}       \\
Ant-mr                     & medium-expert               & 36.5±5.9                & 36.0±6.7                 & 19.1±1.6                 & 36.2±6.6                 & 37.2±4.7                 & 54.2±5.2                 & \textbf{58.8±3.6}       \\
Ant-mr                     & expert                      & 24.4±4.8                & 22.1±0.4                 & 19.5±0.8                 & 27.1±3.7                 & 24.3±2.8                 & \textbf{74.7±10.5}       & 43.8±2.6                \\
Half-m                     & medium                      & 12.3±1.2                & 10.6±1.2                 & 8.3±1.2                  & 16.8±4.2                 & 23.6±5.7                 & \textbf{40.2±0.0}        & 38.5±1.4                \\
Half-m                     & medium-expert               & 10.8±1.9                & 12.9±2.8                 & 8.7±1.3                  & 10.3±2.7                 & 9.8±2.4                  & 10.1±4.0                 & \textbf{19.1±1.0}       \\
Half-m                     & expert                      & 12.6±1.7                & 12.1±1.0                 & 10.8±1.7                 & 12.2±0.9                 & 12.8±0.7                 & 8.7±2.0                  & \textbf{13.1±0.8}       \\
Half-me                    & medium                      & 21.8±6.5                & 25.9±7.4                 & 30.0±4.3                 & 17.2±3.3                 & 21.9±6.5                 & 30.7±9.6                 & \textbf{38.4±1.4}       \\
Half-me                    & medium-expert               & 7.6±1.4                 & 9.5±4.2                  & 6.8±2.9                  & 9.6±2.4                  & 8.9±3.3                  & 10.9±4.2                 & \textbf{24.1±4.6}       \\
Half-me                    & expert                      & 9.1±2.4                 & 10.4±1.3                 & 4.9±3.2                  & 11.2±1.0                 & 10.7±1.4                 & 3.2±0.6                  & \textbf{13.4±2.0}       \\
Half-mr                    & medium                      & 10.0±5.4                & 11.5±4.9                 & 7.5±3.1                  & 10.2±3.7                 & 11.6±4.6                 & \textbf{37.8±2.1}        & 19.5±1.8                \\
Half-mr                    & medium-expert               & 6.5±3.1                 & 9.2±4.7                  & 6.6±1.7                  & 9.5±1.8                  & 8.6±2.3                  & 9.7±2.0                  & \textbf{11.4±2.1}       \\
Half-mr                    & expert                      & 13.6±1.4                & 14.8±2.0                 & 10.4±4.9                 & 14.8±2.2                 & 13.9±2.2                 & 7.2±1.4                  & \textbf{15.6±2.9}       \\
Hopp-m                     & medium                      & 58.7±8.4                & 43.9±15.2                & 12.3±6.6                 & 65.4±1.5                 & 65.3±1.4                 & 65.6±1.9                 & \textbf{69.8±2.3}       \\
Hopp-m                     & medium-expert               & 68.5±12.4               & 55.4±16.9                & 15.6±10.8                & 43.9±30.8                & 51.1±18.5                & 55.4±25.1                & \textbf{78.2±5.1}       \\
Hopp-m                     & expert                      & 79.9±35.5               & 83.7±19.6                & 14.8±5.5                 & 53.1±39.8                & \textbf{87.4±25.4}       & 35.0±19.4                & 59.8±21.8               \\
Hopp-me                    & medium                      & 66.0±0.5                & 61.1±4.0                 & 35.0±20.1                & 64.6±2.6                 & 65.2±1.5                 & 65.3±2.4                 & \textbf{69.6±1.3}       \\
Hopp-me                    & medium-expert               & 45.1±15.7               & 61.9±16.9                & 13.9±4.9                 & 54.7±17.0                & 62.9±15.6                & 38.6±15.9                & \textbf{75.5±9.6}       \\
Hopp-me                    & expert                      & 44.9±19.8               & 84.2±21.1                & 12.0±4.3                 & 57.6±40.6                & 52.8±19.7                & 29.9±11.3                & \textbf{64.5±24.2}      \\
Hopp-mr                    & medium                      & 36.0±0.1                & 39.4±7.2                 & 3.2±2.6                  & 36.1±0.2                 & 35.9±2.4                 & 35.5±12.2                & \textbf{64.8±2.4}       \\
Hopp-mr                    & medium-expert               & 36.1±0.1                & 34.1±3.6                 & 4.4±2.8                  & 36.0±0.1                 & 36.1±0.1                 & 47.5±14.6                & \textbf{69.7±7.5}       \\
Hopp-mr                    & expert                      & 36.0±0.1                & 36.1±0.2                 & 3.7±2.5                  & 36.1±0.1                 & 36.1±0.3                 & 49.9±30.5                & \textbf{69.9±18.0}      \\
Walker-m                   & medium                      & 34.3±9.8                & 35.2±22.5                & 14.3±11.2                & 39.0±6.7                 & 41.9±11.2                & 49.6±18.0                & \textbf{63.2±4.2}       \\
Walker-m                   & medium-expert               & 30.2±12.5               & 51.9±11.5                & 13.6±7.7                 & 38.6±6.5                 & 42.3±19.3                & 43.5±16.4                & \textbf{53.5±7.0}       \\
Walker-m                   & expert                      & 56.4±18.2               & 40.7±14.4                & 15.3±2.5                 & 57.3±12.2                & 60.4±17.5                & 46.7±13.6                & \textbf{70.5±12.0}      \\
Walker-me                  & medium                      & 41.8±8.8                & 38.1±14.4                & 21.4±8.3                 & 36.9±4.3                 & 41.2±13.0                & 44.6±6.0                 & \textbf{59.4±6.8}       \\
Walker-me                  & medium-expert               & 22.2±8.7                & 23.6±8.1                 & 15.9±4.1                 & 23.2±7.9                 & 28.1±4.0                 & 16.5±7.2                 & \textbf{53.2±7.3}       \\
Walker-me                  & expert                      & 26.3±10.4               & 36.0±9.2                 & 18.5±3.6                 & 40.9±9.6                 & 46.2±19.4                & 42.4±9.1                 & \textbf{69.2±7.0}       \\
Walker-mr                  & medium                      & 11.5±7.1                & 12.5±4.3                 & 1.9±2.1                  & 14.3±3.1                 & 22.2±5.2                 & 49.7±9.7                 & \textbf{52.9±8.4}       \\
Walker-mr                  & medium-expert               & 9.7±3.8                 & 11.2±5.0                 & 4.6±3.0                  & 4.2±5.1                  & 7.6±4.9                  & \textbf{55.9±17.1}       & 36.4±5.4                \\
Walker-mr                  & expert                      & 7.7±4.8                 & 7.4±2.4                  & 3.6±1.5                  & 13.2±8.5                 & 7.5±2.1                  & \textbf{51.9±7.9}        & 44.4±8.5                \\ \midrule
Total Score                &                             & 1193.0                    & 1219.8                   & 513.5                    & 1195.7                   & 1271.0                     & 1547.6                   & \textbf{1902.2}         \\ \bottomrule
\end{tabular}
\end{table*}

\subsection{Effect of Guided-Diffusion Model}
In this section, we provide additional experiment results to answer question 3. Specifically, we compare DmC against three variants: (1) DmC without upsampling via the diffusion model (\emph{w/o diffusion}), (2) DmC where the target dataset is upsampled using a diffusion model (\emph{upsample target}), and (3) DmC with naive diffusion-based generation without guidance (\emph{w/o guidance}). We conduct experiments on Ant \emph{medium} and \emph{medium-expert} source datasets under gravity shift and present results in Table \ref{tab:appendix_diffusion_effect}.
We observe that \emph{upsample target} slightly improves performance over \emph{w/o diffusion}. However, the \emph{w/o guidance} variant, which naively upsamples the source dataset, achieves significantly better results than \emph{upsample target}. We hypothesize that the limited target data setting hinders the training diffusion model in the target domain, thus affecting adaptation performance. Among all variants, DmC achieves the highest performance, demonstrating the effectiveness of our proposed method of using a guided-diffusion model to upsample the source dataset with generated data close to the target domain.

\begin{table}
\caption{Performance comparison between DmC and its variants on using the diffusion model. We bold the highest scores.m=medium, e=expert, me=medium-expert.}
\label{tab:appendix_diffusion_effect}
\centering
\begin{tabular}{@{}llll@{}}
\toprule
Method           & m-m               & m-me              & m-e                \\ \midrule
w/o diffusion    & 23.6±2.9          & 14.4±0.8          & 19.7±1.4           \\
up sample target & 25.4±3.1          & 23.7±3.7          & 24.2±2.7           \\
w/o guidance     & 46.8±11.5         & 31.6±11.5         & 22.43.6            \\
w guidance (DmC)      & \textbf{56.9±2.2} & \textbf{47.5±3.9} & \textbf{36.1±7.8}  \\ \midrule
Method           & me-m              & me-me             & me-e               \\ \midrule
w/o diffusion    & 22.5±3.7          & 16.1±1.3          & 18.4±1.7           \\
up sample target & 22.6±2.8          & 27.1±1.1          & 22.6±2.6           \\
w/o guidance     & 40.7±6.5          & 35.8±7.2          & 30.2±4.1           \\
w guidance (DmC)      & \textbf{55.7±8.1} & \textbf{46.3±4.9} & \textbf{42.7±13.0} \\ \bottomrule
\end{tabular}
\end{table}

\subsection{Guided Score $\kappa$}
The parameter $\kappa$ controls the conditional $k$-NN score used for generating source samples, effectively determining their proximity to the target domain. To evaluate its impact, we conduct experiments on HalfCheetah and Hopper under both gravity and kinematic shifts. The results, presented in Table \ref{tab:guided_score}, show that a lower $\kappa$ value ($\kappa = 80$) leads to significant performance drops across multiple tasks. We hypothesize that this occurs because the generated source samples are not sufficiently beneficial for policy adaptation. These findings highlight the importance of generating source samples that closely align with the target domain to improve adaptation performance.

\begin{table}[]
\centering
\caption{Performance comparison of DmC between different guided score $\kappa$ values. We highlight the highest score in bold. Half=Halfcheetah, Hopp=Hopper, m=medium, me=medium-expert, e=expert.}
\label{tab:guided_score}


\begin{tabular}{@{}lllll@{}}
\toprule
\multirow{2}{*}{Source} & \multirow{2}{*}{Target} & \multicolumn{3}{c}{$\kappa$}                               \\ \cmidrule(l){3-5} 
                        &                         & 80                & 90                 & 99                \\ \midrule
Half-gravity-m          & me                      & 48.2±0.8          & \textbf{48.9±0.7}  & 47.8±1.8          \\
Half-gravity-m          & e                       & 48.8±0.4          & 48.8±1             & \textbf{49.5±1}   \\
Half-kinematic-m        & me                      & 17.7±4.1          & \textbf{19.1±1}    & 16.7±3.6          \\
Half-kinematic-m        & e                       & \textbf{13.7±1.3} & 13.1±0.8           & 13.2±1.4          \\
Hopp-gravity-m          & me                      & 33.6±8            & \textbf{40.6±5.5}  & 35±4.9            \\
Hopp-gravity-m          & e                       & 32.8±8.5          & \textbf{51.3±10.6} & 40.5±12.2         \\
Hopp-kinematic-m        & me                      & 74.6±4.2          & 78.2±5.1           & \textbf{79.3±8.5} \\
Hopp-kinematic-m        & e                       & 53.4±20.6         & \textbf{59.8±21.8} & 53.5±18.2         \\ \bottomrule
\end{tabular}
\end{table}

\subsection{Effect of K-NN estimator}
Table~\ref{tab:knn_only} presents the overall performance scores of DmC-K and the baselines on target tasks (with detailed results provided in Appendix Table 8). As shown, DmC-K consistently outperforms methods that rely on neural networks to estimate domain gaps, achieving performance improvements ranging from 50\% to 144\%. Additionally, despite its simplicity, DmC-K performs on par with, or slightly better than, OTDF. These results highlight the effectiveness of the $k$-NN estimator in accurately capturing the dynamics gap between the source and target domains.
\begin{table}[h]
\centering
\caption{Performance comparison between DmC-K and cross-domain RL baselines utilizing NN-based domain gap estimation. We bold the highest scores and underline the second-best. m=medium, e=expert, me=medium-expert, mr=medium-replay.}
\label{tab:knn_only}
\begin{tabular}{lccccc}
\hline
\textbf{Task} & \textbf{DARA} & \textbf{IGDF} & \textbf{SRPO}      & \textbf{OTDF}         & \textbf{DmC-K}         \\ \hline
Ant me-m      & 8.1$\pm$1.8   & 8.9$\pm$1.5   & 8.4$\pm$2.1        & 18.6$\pm$11.9         & \textbf{27.6$\pm$6.0}  \\
Ant me-me     & 6.4$\pm$1.4   & 7.2$\pm$2.9   & 6.1$\pm$3.5        & \textbf{34.0$\pm$9.4} & {\ul 15.4$\pm$1.0}     \\
Ant me-e      & 10.4$\pm$2.9  & 9.2$\pm$1.5   & 8.8$\pm$1.0        & \textbf{23.2$\pm$2.9} & {\ul 18.0$\pm$1.6}     \\
Ant mr-m      & 21.7$\pm$2.1  & 19.6$\pm$1.0  & 18.7$\pm$1.7       & {\ul 29.6$\pm$10.7}   & \textbf{31.7$\pm$0.7}  \\
Ant mr-me     & 18.3$\pm$2.1  & 20.3$\pm$1.6  & 18.7$\pm$1.8       & {\ul 25.4$\pm$2.1}    & \textbf{26.9$\pm$1.1}  \\
Ant mr-e      & 20.0$\pm$1.3  & 18.8$\pm$2.1  & 19.9$\pm$2.1       & {\ul 24.5$\pm$2.8}    & \textbf{26.0$\pm$0.61} \\
Walker me-m   & 41.6$\pm$13.0 & 33.8$\pm$3.1  & {\ul 46.8$\pm$3.5} & 30.2$\pm$9.8          & \textbf{56.1$\pm$4.4}  \\
Walker me-me  & 45.8$\pm$9.4  & 44.7$\pm$2.9  & 36.4$\pm$3.4       & {\ul 53.3$\pm$7.1}    & \textbf{63.9$\pm$1.4}  \\
Walker me-e   & 56.4$\pm$3.5  & 45.3$\pm$10.4 & 45.8$\pm$8.0       & {\ul 61.1$\pm$3.4}    & \textbf{67.0$\pm$2.3}  \\
Walker mr-m   & 14.1$\pm$6.1  & 11.6$\pm$4.6  & 17.9$\pm$3.8       & \textbf{32.7$\pm$7.0} & {\ul 31.8$\pm$8.6}     \\
Walker mr-me  & 15.9$\pm$5.8  & 13.9$\pm$6.5  & 15.3$\pm$4.5       & \textbf{31.6$\pm$6.1} & {\ul 27.5$\pm$3.5}     \\
Walker mr-e   & 15.7$\pm$4.5  & 15.2$\pm$5.3  & 13.7$\pm$8.1       & {\ul 31.3$\pm$5.3}    & \textbf{34.0$\pm$2.8}  \\ \hline
\end{tabular}
\end{table}

\begin{figure}[!ht]
    \centering
    \includegraphics[width=0.8\columnwidth]{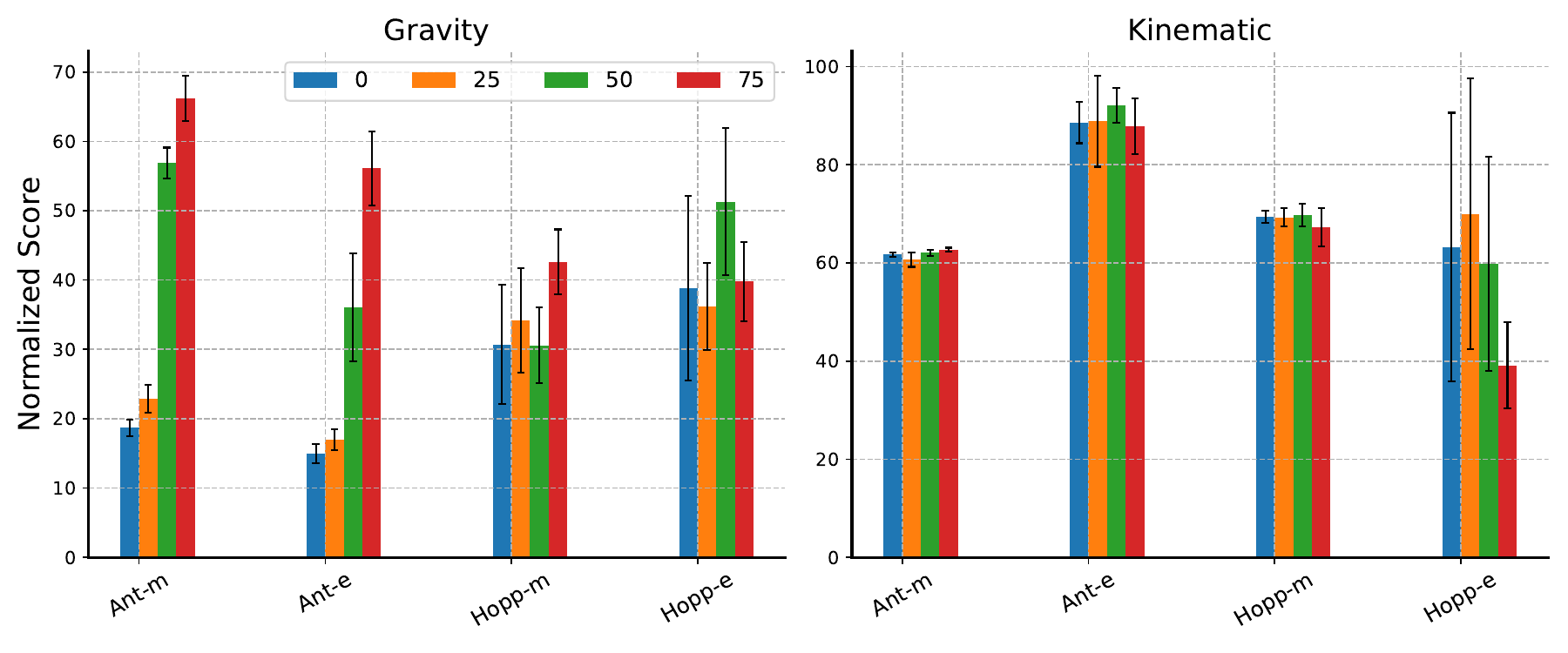}
    \caption{Parameter study of the source data selection ratio. We report the results of Ant and Hopper tasks.}
    \label{fig:appendix_cut_off}
\end{figure}
\bigskip

\subsection{Effect of $\lambda$}
\begin{figure*}[]
    \centering
    \includegraphics[width=0.9\linewidth]{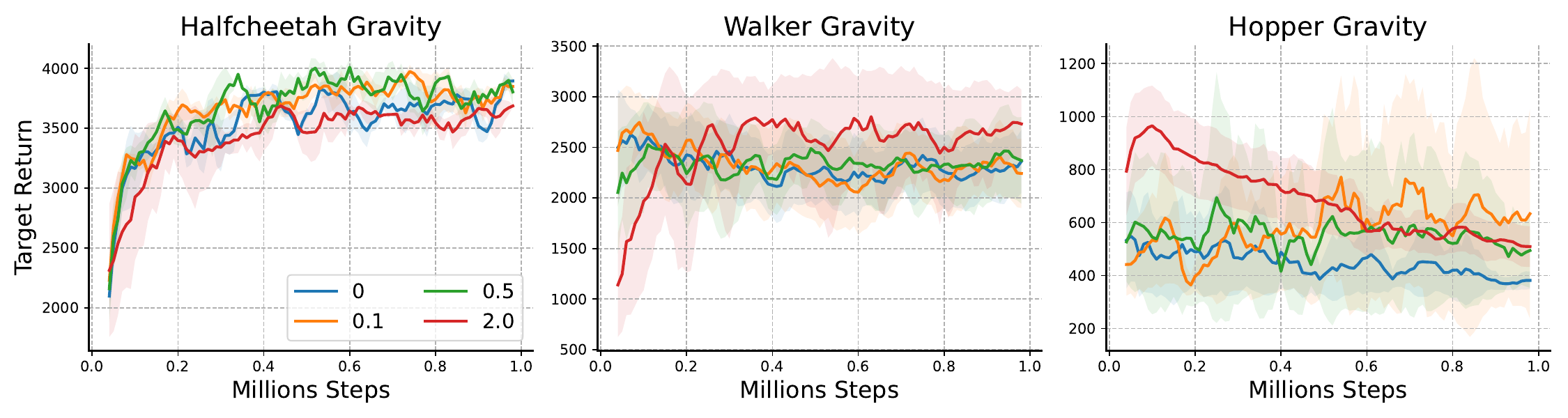}
    \caption{Ablation study on the effect of policy regularization weight $\lambda$. The solid curves are the average target returns, and the shaded areas represent the standard deviation over 5 random seeds.}
    \label{fig:appendix_lambda}
\end{figure*}

$\lambda$ controls the strength of the policy regularization in Eq (11) in the main paper. A small $\lambda$ may lead to the policy being biased toward the source dataset, while a large $\lambda$ limits knowledge transfer from the source. We evaluate different $\lambda$ values ($\lambda \in {0, 0.1, 0.5, 2}$) and present the results in Figure \ref{fig:appendix_lambda}. Removing policy regularization, i.e. setting $\lambda=0$, leads to suboptimal performance. We also observe that the optimal $\lambda$ varies by task (e.g., Halfcheetah prefers $\lambda=0.1$, while Walker performs best with $\lambda=2.0$). To balance this trade-off, we set $\lambda = 0.1$ for all tasks.

\subsection{Data Selection Ratio $\xi$}
Parameter $\xi$ controls how many source data we select to use in a batch at each training step for policy learning, with larger $\xi$ indicating more source data will be rejected. We evaluate the impact of $\xi$ on DmC's performance using medium source datasets under both gravity and kinematic shifts. Specifically, we sweep $\xi \in {0, 25, 50, 75}$, where $\xi=0$ means all source data is used for training. As shown in Figure \ref{fig:appendix_cut_off}, using $\xi=0$ is suboptimal, confirming that naively combining source and target datasets is ineffective. While different tasks prefer different $\xi$ values, we find that $\xi=50$ provides a balanced trade-off across tasks.

\subsection{Effect of Weighting Q-function}
We conduct experiments to evaluate the impact of weighting the Q-function with the score as in Eq. (10) in the main paper. Specifically, we compare DmC with a variant that removes this weighting during value function updates, as follows:
\begin{equation}\label{eq:appendix_wo_weight_q}
    \begin{aligned}
        \mathcal{L}_{Q} = \mathbb{E}_{D_{tar}}\left[(Q_\phi - \mathcal{T}Q_\phi)^2\right] 
        + \mathbb{E}_{D_{src}}\left[\mathds{1}(w_k \geq w_{k,\xi\%})(Q_\phi - \mathcal{T}Q_\phi)^2\right],
    \end{aligned}
\end{equation}
Since the source data selection ratio remains constant, Eq. (\ref{eq:appendix_wo_weight_q}) may suffer from bad source transitions, as low-quality source samples can still be used for training. Weighting the Q-function mitigates this issue by reducing the influence of source samples that deviate significantly from the target domain. As shown in Table \ref{tab:appendix_weight_q}, removing the weighting mechanism leads to performance drops in 3 out of 4 tasks, highlighting its importance.
\begin{table}[ht] 
\centering
\caption{Ablation study on the effect of weighting q-function. We report the normalized score on the target domain. The highest scores are bold.}
\label{tab:appendix_weight_q}
\begin{tabular}{@{}lll@{}}
\toprule
Task                & \multicolumn{1}{c}{w/o weighting q} & \multicolumn{1}{c}{w weighting q} \\ \midrule
Ant-kinematic       & 87.5±6.6                            & \textbf{92.1±3.5}                 \\
Hopper-kinematic    & 51.1±18.9                           & \textbf{59.8±21.8}                \\
Halfcheetah-gravity & \textbf{49.4±0.8}                   & 48.8±1                            \\
Walker-gravity      & 62.4±1.9                            & \textbf{63.8±2.7}                 \\ \bottomrule
\end{tabular}
\end{table}

\end{document}